\documentclass[conference]{IEEEtran}
\IEEEoverridecommandlockouts
\usepackage{mathptmx} % This is Times font
\usepackage{fancyhdr}
\usepackage{verbatim}
\usepackage{amsmath}
\usepackage[normalem]{ulem}
\usepackage[hyphens]{url}
\usepackage[sort,nocompress]{cite}
\usepackage[final]{microtype}
\usepackage[keeplastbox]{flushend}
\usepackage[bookmarks=true,breaklinks=true,letterpaper=true,colorlinks,linkcolor=black,citecolor=blue,urlcolor=black]{hyperref}

\usepackage{multirow}
\usepackage{diagbox}
\usepackage{booktabs}
\usepackage{xcolor} 
\usepackage{xspace}
\usepackage{balance}
\usepackage{soul}

\usepackage{xspace}
\usepackage{balance}
\usepackage{soul}

\usepackage{marginnote}

\usepackage{subcaption}
\usepackage{hyperref}

\definecolor{olivegreen}{rgb}{0, 0.6, 0}
\newcommand{\JL}[1]{{\color{magenta}[\textbf{\sc JLee}: \textit{#1}]}}
\newcommand{\JLr}[1]{{\color{blue}[\textbf{\sc JLee}: \textit{#1}]}}
\newcommand{\SR}[1]{{\color{blue}[\textbf{\sc SRyu}: \textit{#1}]}}

\newcommand{\num}[1]{{\color{magenta}[\textit{#1}]}}
\newcommand{\pimname}{GradPIM\xspace}
\renewcommand{\JL}[1]{}
\renewcommand{\JLr}[1]{}
\renewcommand{\SR}[1]{}
\renewcommand{\num}[1]{#1}

\newcommand{\rindex}[1]{\label{rev:#1}\marginnote{{\color{magenta}\textbf{#1}}}}
\renewcommand{\rindex}[1]{}
\newcommand{\revref}[1]{\hyperref[rev:#1]{\color{magenta}#1}}
\newcommand{\figref}[1]{\hyperref[fig:#1]{\color{blue}\figurename~\ref{fig:#1}}}
\newcommand{\rev}[1]{{\color{olivegreen}#1}}
\renewcommand{\rev}[1]{#1}
\usepackage{tikz}
\newcommand*\circled[1]{\tikz[baseline=(char.base)]{
            \node[shape=circle,draw,inner sep=0.4pt] (char) {#1};}}

\newcommand\blfootnote[1]{%
  \begingroup
  \renewcommand\thefootnote{}\footnote{#1}%
  \addtocounter{footnote}{-1}%
  \endgroup
}

\begin{document}

\title{GradPIM: A Practical Processing-in-DRAM Architecture for Gradient Descent}

\author{\IEEEauthorblockN{Heesu Kim\textsuperscript{1,2}, Hanmin Park\textsuperscript{3}, Taehyun Kim\textsuperscript{1}, Kwanheum Cho\textsuperscript{5}, Eojin Lee\textsuperscript{4,$\dagger$},\\Soojung Ryu\textsuperscript{1,2}, Hyuk-Jae Lee\textsuperscript{1}, Kiyoung Choi\textsuperscript{1,2}, and Jinho Lee\textsuperscript{5,*}}
\IEEEauthorblockA{
\textit{\textsuperscript{1}ECE, Seoul National University\ \ \ \ \ \ }
\textit{\textsuperscript{2}NPRC, Seoul National University} \\
\textit{\textsuperscript{3}SAIT, Samsung Electronics\ \ \ \ }
\textit{\textsuperscript{4}Memory Business, Samsung Electronics\ \ \ \ }
\textit{\textsuperscript{5}CS, Yonsei University} \\
}
\IEEEauthorblockA{
\textit{muncok@dal.snu.ac.kr, hanmin.park@samsung.com, kimth@capp.snu.ac.kr, gh350@yonsei.ac.kr, } \\
\textit{eojin29.lee@samsung.com, s.ryu@snu.ac.kr, hjlee@capp.snu.ac.kr, kchoi@snu.ac.kr, leejinho@yonsei.ac.kr}
}
}

\maketitle

\begin{abstract}
In this paper, we present \pimname, a processing-in-memory architecture which accelerates parameter updates of deep neural networks training. 
As one of processing-in-memory techniques that could be realized in the near future, we propose an incremental, simple architectural design that does not invade the existing memory protocol.
Extending DDR4 SDRAM to utilize bank-group parallelism makes our operation designs in processing-in-memory (PIM) module efficient in terms of hardware cost and performance.
Our experimental results show that the proposed architecture can improve the performance of DNN training and greatly reduce memory bandwidth requirement while posing only a minimal amount of overhead to the protocol and DRAM area. 
% \begin{abstract}
% In this paper, we present \pimname, a processing-in-memory architecture which accelerates the parameter updates of deep neural networks training. 
% Towards processing-in-memory techniques that could be realized in the near future, we propose designing an incremental, simple architecture that does not invade the existing protocol.
% Simple extension of DDR4 SDRAM to utilize bank-group parallelism makes our operation designs in processing-in-memory (PIM) module efficient in terms of hardware cost and performance.
% Our experimental results show that the proposed architecture can improve the performance of DNN training and greatly reduce memory bandwidth requirement while posing only a minimal amount of overhead to the protocol and the DRAM area. 
\blfootnote{Author preprint. To appear at HPCA 2021}
\blfootnote{\textsuperscript{$\dagger$} This work was done when H. Park and E. Lee were at Seoul National University.}
\blfootnote{\textsuperscript{*} Correspondence}
\end{abstract}

\vspace{-2mm}
\section{Introduction}
%Training a deep neural network (DNN) is a time-consuming process.  
%For instance, training the widely-used ResNet-50~\cite{resnet} on ImageNet dataset~\cite{imagenet} requires about a week of GPU time for training.
%Building a specialized neural processing unit (NPU) is a promising approach for achieving both speedup and energy efficiency.
As DNN emerges as one of the most significant applications of the era, neural processing units (NPU) are quickly becoming an important member of computing systems for both of inference and training~\cite{eyeriss, eyeriss2, shidiannao, diannao, dadiannao, tpu, scnn}.
%\SR{we may need to mention that it still maintains the 16-bit parameter update.}
Often, the key to achieving the NPU performance is minimizing its memory (usually DRAM) bandwidth requirements. 
Naively executing DNNs would incur heavy memory traffic caused by repeated memory accesses on the same data. 
%
%In addition to simply relying on the mixed-precision techniques for the speedup, we examine various optimization techniques that can be used together, mainly towards minimizing the memory bandwidth consumption of the NPUs. 
%Hence, numerous techniques have been designed towards reducing the traffic to the memory. % revive this if more space
Many NPU architectures\cite{eyeriss, shidiannao, ws1, ws2, diannao, dadiannao} utilize different dataflow models in order to handle the data reuse problem by loop reordering and careful address mapping to the PE array. 
Also, mixed-precision training~\cite{baidu, ibm1, ibm2, intel1, intel2} is a widely used technique that helps reduce both the computation and memory traffic. % \HK{by squeezing the redundant data precision}.
In addition, minibatch serialization~\cite{mbs}, BN fission and fusion~\cite{bnff}, and layer fusion~\cite{pyramid} fall into the class of inter-layer optimizations that finds opportunities of data reuse %by applying inter-layer optimizations 
to reduce the amount of memory traffic. 

\begin{figure}
    \centering
    \includegraphics[width=.95\linewidth]{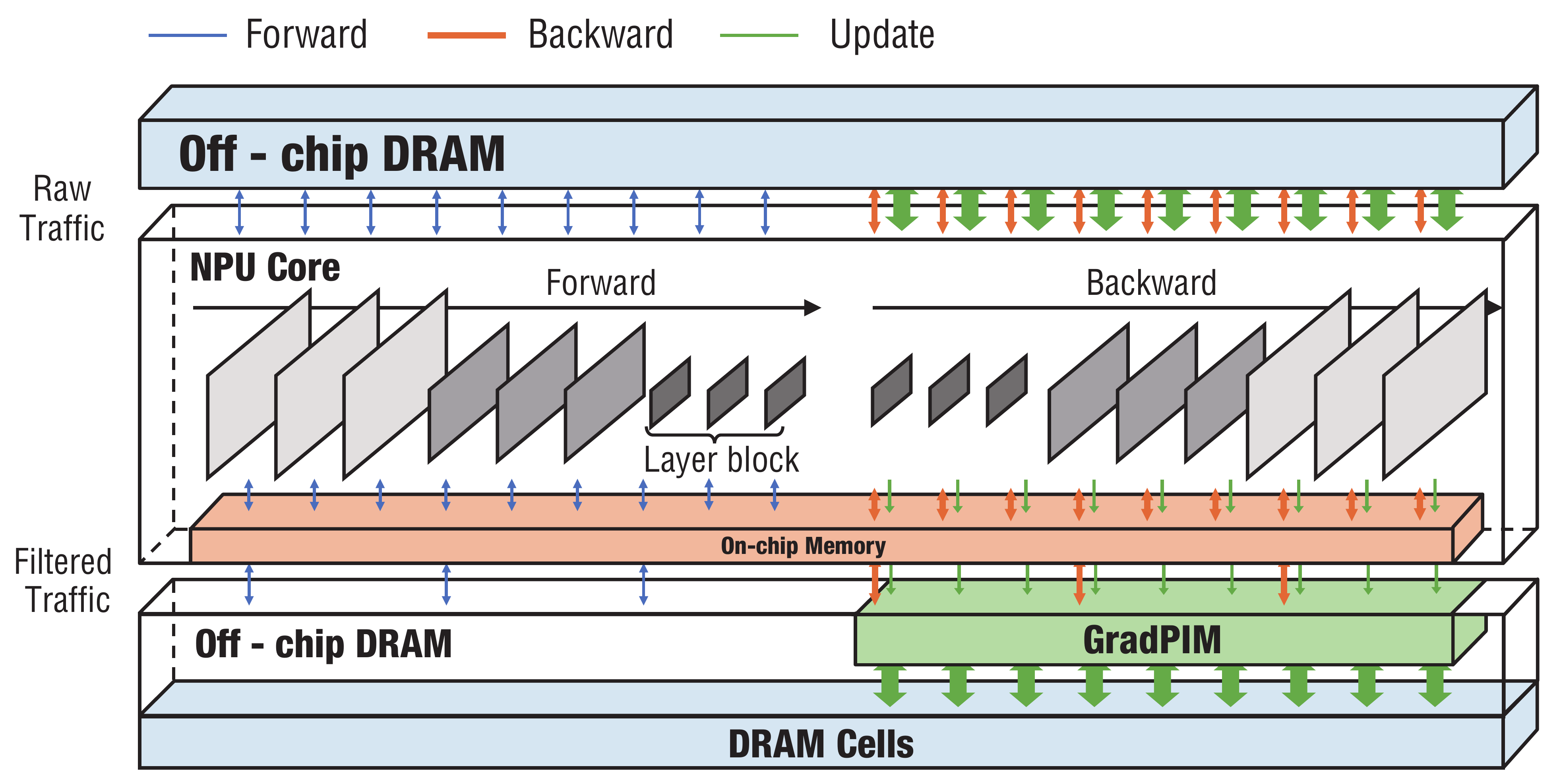}
    \caption{Mixed-precision training with \pimname.}
    \vspace{-7mm}
    \label{fig:intro}
\end{figure}

Processing In Memory (PIM) is a surging technique that optimizes memory accesses. 
By placing logic components and DRAM on the same die (or same stack~\cite{hmc}), the off-chip bandwidth bottleneck can be significantly mitigated as demonstrated by many PIM works~\cite{neurocube, drisa, dracc, google}. 
However, despite its huge potential, bringing PIM to the market has always been a tough challenge. 
%\label{rev:R6a}
\rindex{R6a}\rev{One of the major reasons was that making the necessary changes (e.g., ISA extension, memory controller support, etc.) from the CPU side is too disruptive to the current ecosystem. 
However, we believe it could be far easier to do so on NPUs, as its ecosystem is not fixed yet, and there are numerous vendors trying to design their own NPUs. 
%Considering the rapid growth and the non-mature-yet environment of NPUs, 
In such circumstances, NPUs can become a major opportunity to bringing PIMs to the DNN execution ecosystem.} %deleted "we believe"

Recently, Kim~\cite{practical} revealed some practical challenges from the DRAM vendors' view on what is hindering commercial DRAM products from supporting PIM. 
He discussed issues such as 3D stacking overhead, limited power/thermal budget, and conflicts with existing protocols.
From a similar perspective, we predict that if a PIM technique were to be accepted by commercial products in the near future, it would be incremental to the existing products (e.g., DDR, HBM).
In that sense, a practical PIM solution for DNN training should achieve the following three goals: 
\begin{enumerate}    
    \item \textbf{Fixed-function PIM~\cite{taxonomy}, non-invasive to the existing protocol:} The technique should preserve the existing protocol as much as possible. The new functions should have deterministic latencies in order to prevent conflict with the current memory controller designs. 
    \item \textbf{Simple, memory-intensive functions:} To reduce the area overhead and mitigate the thermal/power budget problem while allowing for a large performance gain, the added function should be simple, and memory-intensive. 
    \item \textbf{Isolation of PIM components and DRAM cell array:} Modifying the cell array would cause radical design changes to the existing product, and should be avoided.
\end{enumerate}

Taking the above goals into careful consideration, we propose \pimname, a fixed-function PIM that supports gradient descent logic within the DRAM die. 
%According to our analysis of the state-of-the-art data reuse techniques, we found that the parameter update phase of DNN training is an appropriate target for a cost-effective PIM. 
%For example, in ResNet-18~\cite{resnet}, %from reading and writing high precision master weights and the associated parameters, 
%the parameter update phase consumes up to about 50\% of the memory bandwidth and\JL{@@} the execution time.
According to our analysis of state-of-the-art data reuse techniques, parameter updates of modern DNN workloads have high memory bandwidth requirements. % caused by accessing high-precision data. 
For example, that of ResNet-18~\cite{resnet} consumes up to 50\% of the total memory traffic during training. 
This memory-intensive nature of parameter update makes PIM an appropriate solution.
%Moreover, even though the computational intensity is very low, there has to be a dedicated high precision datapath for the high-precision parameter updates in the NPUs.
%Keeping the master weight in a fixed-point is one way to reduce the hardware cost~\cite{fxpnet}, but becomes a burden to the NPU in both processing and transferring of high-precision data.

Unlike the other parts of the training, the operations involved in the update phase have been overlooked in many designs. 
Those operations are relatively simple, and do not have much room for further optimization. 
However, being both simple and memory-intensive makes them an excellent target for processing-in-memory. 
In this paper, we propose assisting a mixed-precision accelerator with a PIM-enabled memory specialized for parameter update operations. 

Conceptually, we attempt to isolate the memory traffic associated with the parameter update phase within the memory using \pimname (\figurename~\ref{fig:intro}), just as data reuse techniques isolate the traffics of forward and backward operations within NPUs. 
We identify the bank group I/O gating as the ideal place for placing the parameter update logic.
By placing registers next to the bank group I/O gating, we effectively decouple each bank group from the global DRAM structure. 
With careful data arrangements, we implement a set of DDR-based PIM operations that can perform parameter updates using the DRAM internal parallelism.

\pimname is designed to be a simple extension from the existing DDR4 protocol~\cite{DDR4} without altering the existing commands. 
The design is also non-invasive to the DRAM cell arrays and places only a small module along with the peripherals.
Those modules are fully controlled by the memory controller using a reserved command (i.e. RFU), and thus \pimname can be considered as a DDR-compatible device rather than an independent accelerator. \JL{better description?}

%With the update unit, we process quantization, dequantization, applying momentum and learning rates, and writing of the master (full-precision) weights. 
%We dedicate certain memory regions only for the certain PIM operations, and injecting \JL{better word than injection?} existing rd/wr commands to those region \JL{is this really possible?. also, maybe apply programmable decoders} automatically translates into pre-configured PIM operations, requiring no modifications to the memory controller IPs.
We evaluate \pimname combined with a contemporary NPU design.
We show that the proposed design significantly reduces the total memory bandwidth requirements and the total execution time.
% We also show that \pimname can also provide speedup toward accelerating the distributed deep learning with multiple NPUs, and can still provide benefits under various configurations, including HMC, HBM, and @@ based designs \JL{can't think of too much right now}.
Our contributions can be summarized as follows:
\begin{itemize}
    \item 
    %For mixed-precision training 
    With modern optimizations towards reducing bandwidth requirements, we identify that the update phase during DNN training accounts for a significant portion, and can be an excellent target for a practical PIM design.
    \item We design a PIM logic that decouples each bank group from the global structures, so that parameter update phase operations can be executed within the DRAM, by utilizing the internal bandwidth of the DRAM.
    \item We model the proposed PIM logic design by performing a layout under DRAM technologies and analyze the area and power overhead of placement.
   \item We address the data placement and mapping issue for deploying \pimname on modern DRAM organizations.
    %\item We build a simulator for both the NPU and PIM design, and evaluate the performance and power gains from the proposed PIM design. \SR {PIM design in RTL?}
\end{itemize}

\begin{figure}
    \centering
    \includegraphics[width=0.9\linewidth]{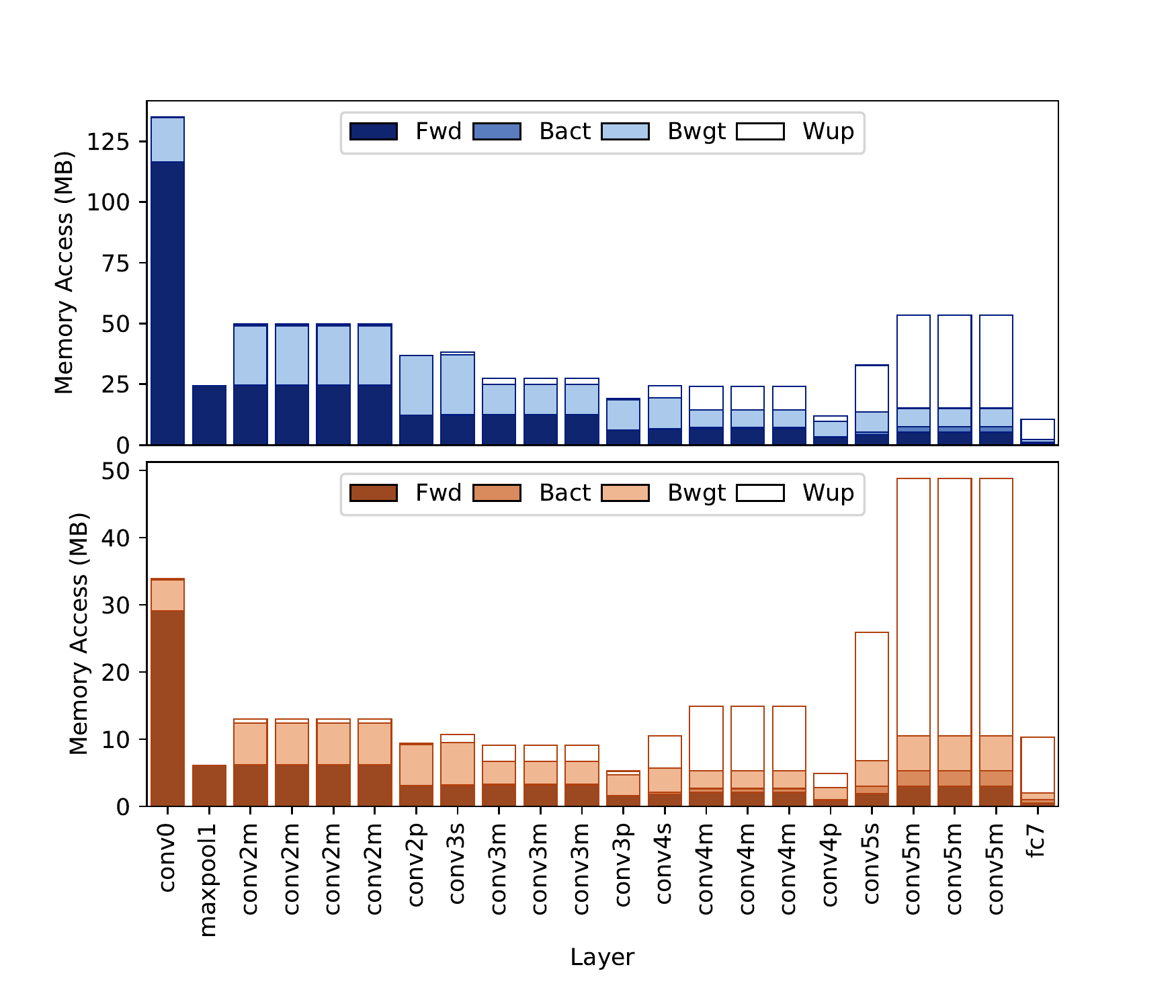}
    \caption{\rev{Breakdown of the memory accesses of ResNet-18 layers for full-precision (top) and mixed-precision (bottom) training. Note that the Y axes have different scales.}}
    \vspace{-4mm}
    \label{fig:analysis}
\end{figure}

\begin{figure*}
% \begin{figure}
    \centering
 \includegraphics[width=.95\textwidth]{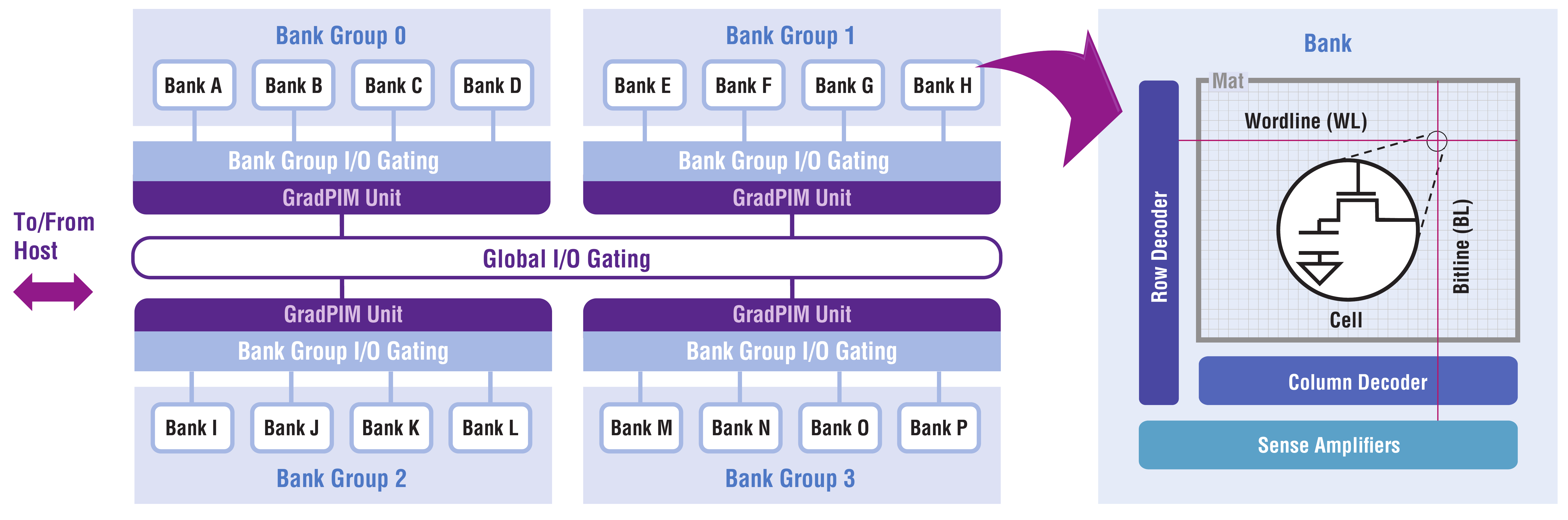}
\caption{Modern DDR4 SDRAM internal architecture.}
    \label{fig:dram}
\end{figure*}
% \end{figure}

\section{Motivation}
\label{sec:moti}
Most NPU designs are focused on maximizing MAC utilization and minimizing memory traffics during forward and backward passes.
Considering the usual training procedures, the traffics from reading activation values from memory dominate the entire memory traffic and the execution time, making the above strategy promising.
However, with mixed-precision training (e.g., 8bit gradient/32bit master weights) and a few state-of-the-art data reuse techniques~\cite{pyramid, mbs, bnff}, the relative portion of memory accesses from parameter updates dramatically increases.
\figurename~\ref{fig:analysis} shows the breakdown of the amount of memory accesses in training a batch of ResNet-18, in MB.
The bars are divided into forward pass (Fwd), backward pass activations (Bact), backward pass weights (Bwgt), and parameter updates (Pup) \rev{for full-precision training (top) and 8bit/32bit mixed-precision training (bottom)}.
To maximize data reuse, we applied MBS (MiniBatch Serialization)~\cite{mbs} and BNFF (Batch-Normalization Fission and Fusion)~\cite{bnff} to reduce the inter-layer data traffic. 

As the network advances to later layers, the portion of the memory traffic caused by parameter updates becomes significant.
\rindex{R1a}\rev{The weight parameter update phase consumes 22.4\% of the total memory accesses during full-precision training.} During mixed-precision training, on the other hand, the update phase occupies a total of 45.9\%. For the last block (a set of conv5m layers), the parameter update phase takes up as much as 80.5\% of memory traffic alone.
This observation is partly in line with the analysis in \cite{google} which reports that offloading packing and quantization of data to a PIM module can bring speedup and energy gain on various applications including TensorFlow Mobile~\cite{tensorflowmobile}.
\rindex{R9}\rev{Even though quantization for training is still premature in that it does not always yield full accuracy, it is an active field of research that includes 16bit~\cite{baidu,intel2} and 8bit~\cite{ibm1, ibm2, intel1} quantizations, and is quickly being adopted by many NPU designs as a promising way of reducing hardware cost~\cite{tpuv2,tensorcore}.}

The portion of traffic in the update phase depends on the ratio of activations and weight parameters.
As the application fields of DNNs expand towards non-CNN workloads such as AlphaGo~\cite{alphagozero} (a DNN-based boardgame player) or multi-layer perceptrons (MLP)~\cite{mlp} (general problems), we have found that the portion of the weight parameters rises especially for those emerging non-vision DNN applications, which indicates great opportunities for the proposed \pimname going forward.
\vspace{2mm}
%\pimname isolates most of the update traffics within the bank group, and the NPU only needs to send the 8-bit scaled gradient to the memory. 
%This leads to a significant amount of execution time and energy reduction, as will be demonstrated in Section~\ref{sec:eval}.

\section{Background}

\JL{I have removed mixed-precision training in the background. maybe I have to change other parts accordingly}
\subsection{DNN Parameter Update Algorithms}

Majority of DNNs rely on a family of stochastic gradient descent (SGD) as the parameter update algorithm. 
In the simplest form of SGD, all parameters in the network are updated towards the negative direction of the gradient at the end of each minibatch execution. 
The update can be formulated as the following:
\vspace{-0.5mm}
\begin{equation}
\theta_{t+1} = \theta_t - \eta g_t
\label{eq:sgd}
\end{equation}
\vspace{-0.5mm}
where $g_t$ is the gradient vector over the current minibatch and $\eta$ is the learning rate.
To gain faster convergence and/or better accuracy, many advanced parameter update algorithms have been proposed. 
For example, SGD with momentum~\cite{momentum} is formulated as below: 
\vspace{-.5mm}
\begin{align}
\label{eq:momentum_vt}
v_t &= \alpha v_{t-1} - \eta g_t \\
\theta_{t+1} &= \theta_t + v_t
\label{eq:momentum_theta}
\end{align}
\vspace{-.5mm}
where $\alpha$ is a momentum decaying factor and $v$ is the momentum. 
The momentum works as a damping factor, and makes the convergence faster. 
If weight decay term $\beta$ is used in addition, Eq~\ref{eq:momentum_vt} becomes
\vspace{-0.5mm}
\begin{align}
\label{eq:decay_vt}
v_t &= \alpha v_{t-1} - \eta(\beta \theta_t + g_t) 
\end{align}
\vspace{-0.5mm}
There are a few more parameter update algorithms worth mentioning such as Adam~\cite{adam}, AdaGrad~\cite{adagrad}, NAG~\cite{nesterov} or RMSprop~\cite{rmsprop}. 
\begin{comment}
For example, RMSprop is formulated by 
\begin{align}
\label{eq:rmsprop}
E[g^2]_t &= \gamma E[g^2]_{t-1} + (1-\gamma)g^2_t \\
\theta_{t+1} &= \theta_t - \frac{\eta}{\sqrt{E[g^2]_t+\epsilon}}g_t
\end{align}
\end{comment}

These algorithms all exhibit element-wise computations and has a relatively low number of computations per element.
However, they usually require higher precision especially due to their small hyperparameters (e.g., $\eta = 0.01$).
Thus, this can easily become the bandwidth bottleneck in the 
%mixed-precision 
DNN training, especially with mixed-precisions.

\subsection{Modern DRAM Architecture}
The modern DRAM architecture is a result of multi-decade effort of increasing cell density and bandwidth at the same time.
\figurename~\ref{fig:dram} shows the modern DDR4 SDRAM architecture. %The proposed \pimname Unit is not shown.  
A DRAM is composed of multiple banks, which are 2D arrays of 1T1C cells.
To increase the area efficiency, a row of cells in a bank share a wordline (WL), which is used to select the row that should be activated.
A vertical set of cells in a bank share a pair of bitlines (BL, BLB) that is used to deliver the data from/to the cells.
When a row is activated, the small charge stored within each cell flows out to the bitline and is caught by a row of sense amplifiers. 
The sense amplifiers restore the cell capacitor to its original value, and this takes tRAS to complete.

%After tRCD time from the start of activation, reading data of the activated row is possible. a few bits (columns) are chosen, and are propagated through the I/O gating and to the off-chip data bus.

To read data from the activated row, a few bits (column) are chosen. 
After tRCD from the beginning of an activation, data can be read from the activated row.
A column is chosen and propagated through the I/O gating and to the off-chip data bus.
Even though each bank can operate independently, the I/O gating circuitry is shared among them, and each column read command occupies the I/O gating for tCCD. 
Therefore a back-to-back column read command has to be spaced with tCCD.
Also, the data occupies the off-chip data bus for tBURST. 
tCCD and tBURST are usually set to be 4 cycles, providing 64 bytes of data in a burst\footnote{In fact, multiple chips co-operate as a rank to provide 64 bytes in a 4 cycle burst. We omit the details for brevity.}.
After tCL of latency from the assertion of a read command, the data burst starts on the data bus.
In the case of a column write, the data flows in the opposite direction. 
The memory controller can place the data on the bus tCWL time units after the write command assertion, and the I/O gating is again occupied after tCCD.

If a different row has to be accessed for read or write commands, the current row has to be deactivated and a new row has to be opened.
However, if a previous read or write command is on-going on a column, the row has to remain open for tRTP or tWR after the previous read/write command respectively, because the row has to provide the data for a read, or has to wait for the data to be restored for a write.

An important concept introduced since DDR4~\cite{DDR4} is \emph{bank groups}. 
To keep up the internal fetch speed with the increasing off-chip data rate, multiple banks (2,4, or 8) form a bank group, and the I/O gating is partitioned into bank group I/O gating and global I/O gating. 
The result is that if consecutive column accesses are asserted to two different bank groups, the accesses only share the global I/O gating, and still can be spaced with 4 cycles (=tCCD\_S) as in previous generations (i.e., DDR3~\cite{DDR3}).
However, if those accesses are to the single bank group, the data occupies both the bank group I/O gating and the global I/O gating, and the two accesses now have to be scheduled with a longer interval (=tCCD\_L) in-between.
Usually, tCCD\_L is 25\% to 100\% longer than tCCD\_S.

With the introduction of bank groups, there are two kinds of opportunities for in-DRAM processing-in-memory (PIM) technologies. % to make use of. 
First, bank-level parallelism exploits the fact that each bank can be independently accessed, and a single bank alone can provide more than half the bandwidth of the off-chip data bus (the ratio depends on tCCD\_S/tCCD\_L). 
Thus, a DDR4 SDRAM chip with 16 banks has more than 8x bank-internal bandwidth, multiplied by the number of ranks per channel. 
On the other hand, each bank group can also provide more than half the bandwidth of the off-chip data bus. 
Therefore there is more than 2x (for DDR4) or 4x (for DDR5) bank group-internal bandwidth, again multiplied by the number of ranks. 
Each bank group has access to multiple banks and therefore provides an opportunity to work with multiple open rows in separate banks, which would not have been possible when working only with bank parallelism.

 \section{\pimname }

Contrary to many PIM work that perform MAC operation within DRAM~\cite{drisa, dracc, mvid}, we leave the MAC operations entirely to the host NPU. 
Instead, \pimname performs the parameter update phase, following the observations from Section~\ref{sec:moti}.
The remainder of this section describes the in-DRAM structure of \pimname and how it is organized.

\subsection{\pimname architecture}
\label{sec:arch}
\begin{figure}
\centering
 \includegraphics[width=0.80\linewidth]{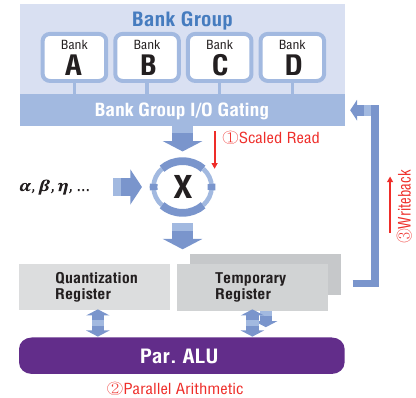}
 \caption{DRAM internal architecture of \pimname unit. \JL{(I can put it  to the left of figure3 for space}}
 \label{fig:drampim}
 \vspace{-3mm}
\end{figure}
  \figurename~\ref{fig:drampim} shows the architecture of \pimname unit. 
  A \pimname unit is placed at each bank group, next to the local I/O gating (\figurename~\ref{fig:dram}). 
  At the heart of \pimname is the temporary registers next to the local I/O gating. 
  They decouple the local I/O gating and the global I/O gating, enabling the bank-group level parallelism~\cite{bglp}. 
  We read to or write from the temporal registers, instead of occupying the external data bus. 
  Also, we perform the vector operations required for the update phase. % in executing DNNs.
  
  Another benefit of placing the \pimname unit next to the bank group I/O gating is that it has access to multiple banks. 
  That means, unlike a few previous PIM approaches~\cite{bufcmp, ambit} which performs operations within a bank, \pimname can operate on multiple rows concurrently at a time.
  This is essential for working with multiple arrays of variables, which would otherwise cost expensive row activation each time a column of different array is accessed.
  
  \pimname logic mainly includes three components: Registers, scaler, and parallel arithmetic unit.
  \begin{itemize}
  \item \emph{Registers} are used to store intermediate results and have the same width of the global sense amplifiers (i.e., 64 Bytes in total for a rank). 
We place two temporary registers per \pimname unit to be used for source and destinations of the arithmetic operations, and one quantization register exclusively for storing the quantized values.
%since this is the most area-expensive part of the \pimname unit\JL{really? chk} and two is the minimum number of registers to perform vector operations.
  
  \item \emph{Scaler} is used to scale the data with pre-defined hyperparameters, e.g., default learning rate. The scaler is placed between the bank group I/O and the registers, and performs element-wise multiplications.

  \item \emph{Parallel arithmetic unit} is used to perform element-wise computations within the update phase, such as the additions in Eq.~\ref{eq:decay_vt}.
  %Therefore, it is composed of multiple ALUs in parallel. 
  In the current version of \pimname, we support simple additions and subtractions.
  
  \end{itemize}

\begin{figure*}
\centering
 \includegraphics[width=0.85\textwidth]{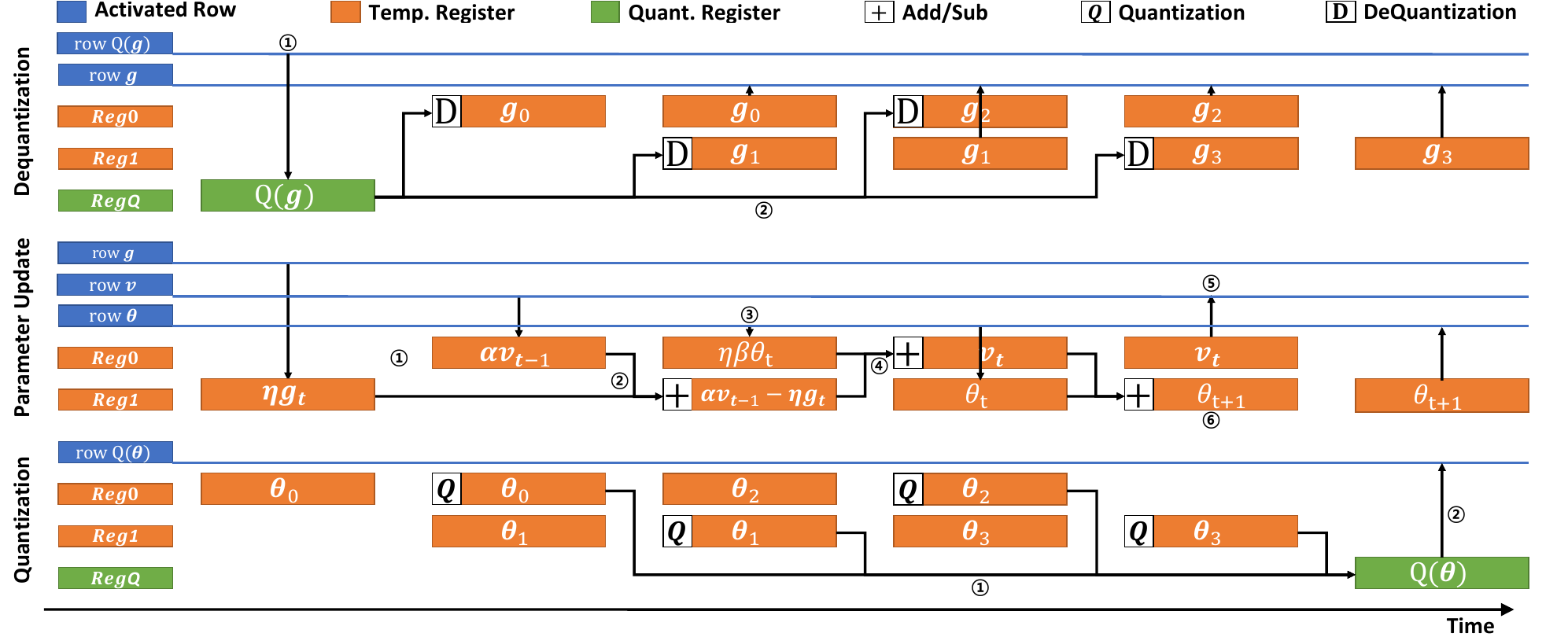}
 \vspace{-2mm}
 \caption{Example procedure for quantization/dequantization and momentum SGD algorithm  with \pimname.}
 \vspace{-2mm}
 \label{fig:procedure}
 \end{figure*}

\subsection{\pimname Operations}
\label{sec:operations}
With the components, the operations that \pimname perform are classified into three categories as below:
\begin{enumerate}

    \item \textbf{Scaled read} loads a column of data into a register from the cells. 
    While loading the registers, the values are scaled by certain hyperparameters, such as $\eta, \alpha$ or $\beta$ as in Eq.~\ref{eq:decay_vt}. 
    Since those values are mostly fixed constants, % that have broadly accepted values in practice, 
    we pin four scaler values to an id to each value.  %(i.e., we assign different opcode for each scaler value). 
    To simplify the scaler, we approximate the scaler values in $2^n \pm 2^m$ and implement the scaler with shifters and adders.
    The values of $n$ and $m$ assigned to each opcode
    %the scalers are implemented using shift-add multiplications, and each register has a designated scaling value to avoid having to provide the scaling value. 
    %Currently, we provide four values values of 1 (no scaling), 0.1, 0.9, and 0.999 that are used in \cite{adam, momentum, rmsprop, adagrad}. 
    %Those constants are stored in mode registers, and 
    can be programmed with MRW (Mode Register Write) command in case the user needs different set of values.

    \item \textbf{Parallel operations} perform arithmetic operations from the registers and puts the result into another register similar to AVX-512 VPADDD~\cite{AVX512} instruction.
    Currently, we support add, sub, quantization, and dequantization to create the partial terms of Eq.~\ref{eq:decay_vt} or execute conversion between quantized and dequantized values.
    Quantization either reads data from one of the temporary registers and write to the quantization register, and vice versa for dequantization.
    Since the quantized values stay longer (four times for 8bit quantizetion) in the register, using a dedicated register greatly simplifies the data and control path circuit design (see Section~\ref{fig:procedure} for details).
  
    \item \textbf{Writeback}. After the operations for the optimizers are complete, the result has to be written back. This corresponds to the second half of the DDR column write, where the register data is written to the global sense amplifier and to the cells.
\end{enumerate}

\subsection{Timing Considerations}
\label{sec:timing}
\JL{seems like we need a bit more to satisfy the DRAM guys}
To allow the memory controller to schedule the \pimname commands along with the existing commands, each \pimname command has to mingle with the timing parameters. We keep timings for each command as below.

    The scaled read is similar to the column read operation in the ordinary DDR protocol~\cite{DDR4}, but the data is placed to one of the registers instead of the data bus and therefore does not limit the scheduling of other commands with tBURST. 
    In an attempt to maintain close consistency with the existing DDR commands, the memory controller regards the operation as complete after tCCD\_L. 
    In DDR protocol, tCCD\_L represents the bandwidth that a bank or a bank group is capable of providing. 
    %tCL is set considering the time for the data to be accessed from the cells to the global I/O. 
    Because the scaled read operation also reads the data from the banks, it is reasonable to assign the same tCCD\_L to read data and be stored in the register. 
    Also, tRTP is still preserved since the sense amplifiers need to provide the data from the cells.
    Please note that the scaled read occupies only the local bank group I/O gating and thus does not interfere with the other scaled read commands in different bank groups.

    The parallel arithmetic operations happen completely out of the conventional DRAM logic and is not governed by the existing timing parameters. To account for the parallel ALU being occupied, we introduce an extra timing parameter tPIM, which represents the worst case execution time for the arithmetic operations.
    This timing parameter does not interfere with any other commands, but prohibits other PIM arithmetic operations from taking place within the same bank group.
    
    Writeback operation can be considered as the latter half of the existing write command. % following the register store. 
    Instead of the data bus, the data comes from one of the registers. Therefore the writeback operation is not affected by tCWL or tBURST, but we keep tCCD\_L as in the scaled read as the bank group I/O gating is occupied. tWR has to comply if the row is to be closed after the writeback since the data propagate into the row through the sense amplifiers.
    
    \JL{this section has some overlap with the prev one. maybe a target for stripping later}
    
     To consider the power budget, we first estimated the maximum power of a DRAM channel as done by \cite{mvid}, by performing sequential reads while keeping the tFAW and tRRD constraints.
     Then we have scaled tFAW and tRRD so that performing consecutive PIM operations %within the new contraints 
     would yield the same maximum power.
     However, it only added a negligible amount of difference to both timing parameters ($<$1\%).
    
\subsection{Update Phase Procedure}
%With \pimname, we perform 3 kinds of phases, which are composed with the operations Section~\ref{sec:operations}.

To execute the update phase with \pimname, the NPU first writes the gradients to the memory generated by a backward pass on the weights. 
Then, the parameter update algorithm is executed as shown in Eq. \ref{eq:momentum_theta} and Eq. \ref{eq:decay_vt}.
After the update, the NPU reads the updated weights for the next step forward pass.
To support mixed-precision training, quantization and dequantization are performed before and after the parameter update algorithm, so that \pimname can work on high-precision data whereas the NPU can work on low-precision data.
In this section, we show how these processes are performed sequentially using the operations described in Section~\ref{sec:operations} assuming 8bit/32bit mixed-precision. In full-precision training, the quantization/dequantization can be omitted. 
%We omit the procedures for the final quantization for space reason since it is similar to dequantization.

\subsubsection{Dequantization}%Because the gradients coming from the NPU are in 8-bit quantized version, they have to be dequantized before applying any calculation to avoid swamping.
%Also, after updating the master weight parameters to avoid having to quantize them each time they are read, \pimname generates a separate 8-bit quantized copy of the weights in the memory.

\figurename~\ref{fig:procedure} (Top) shows the procedure for performing dequantization.
%These are another kind of memory-intensive operation in DNNs, and this is implemented using \pimname as:
We assume the rows for the quantized gradients $Q(g)$ and the dequantized gradients $g$ are already open on different banks within a bank group so that they can be accessed at the same time, and the procedure is as follows:
\circled{1} A column of $Q(g)$ is loaded into the quantize register. \circled{2} A 1/4 of a column of the $Q(g)$ is dequantized and the resulting column is written to a temporary register. 
The dequantzation command specifies which 1/4 of the column should be read from the quantize register and which temporary register to write to (please see Section~\ref{sec:command}).
Then the gradient ($g$) is written back to the corresponding row, and \circled{2} is repeated four times until the entire column had been dequantized.
The procedure is repeated for the consecutive columns of $g$.

One thing to note is that because we place a unit in the local I/O of the bank group, the entire procedure does not experience any row buffer miss except for when a new row is opened for next data accesses (like a cold miss in a cache).

\subsubsection{Parameter Update}
\figurename~\ref{fig:procedure} (middle) shows the procedure for conducting the update phase with \pimname using momentum SGD~\cite{momentum} algorithm as in Eq~\ref{eq:momentum_theta},~\ref{eq:decay_vt} as a simple example. 
We assume that the rows for weight parameters $\theta$, momentum $v$ and gradients $g$ are already activated on different banks within a bank group. % so that they can be open at the same time.
\circled{1}  A column of $g_t$ and $v_{t-1}$ are loaded to the temporary registers, scaled by $\eta$ and $\alpha$ using scaled\_rd operation. 
\circled{2} The scaled values in the two registers above are processed with parallel\_add.
\circled{3} A column of $\theta_t$ is loaded into a temporary register, scaled by $\eta\beta$. 
\circled{4} Parallel\_add is performed once again, creating $v_{t}$ in EQ~\ref{eq:decay_vt}.
\circled{5} Writeback is performed from the register with $v_{t}$ to the open row for $v$.
\circled{6} Similarly, a column of $\theta_{t+1}$ is generated by Eq.~\ref{eq:momentum_theta} and written back to the row storing $\theta$.
Finally, \circled{1} - \circled{6} is repeated for consecutive columns of $g$, $v$ and $\theta$ until the entire row has been processed. 
%If there are still remaining rows to be processed, new rows are activated, and the procedure repeats until all the data have been processed.

This procedure also requires no unnecessary row activations as in the dequantization case.
In the case of the more complicated algorithm where more than one momentum is used per weight parameter, the required number of concurrent open rows might increase, but there are four banks per bank group in typical DDR4/5 SDRAMs and it is enough to cover all per-weight values in most of the SGD-based parameter update algorithms to our knowledge.

\subsubsection{Quantization}

\figurename~\ref{fig:procedure} (Bottom) shows the procedure for performing quantization.
As the last step, the master weight parameters are quantized to a low precision, so that the NPU can read them during the forward and backward phase.
The procedure is similar to dequantization, but the order is opposite.
\circled{1} A column of master weight parameters are loaded to a temporary register, and quantization is performed. It fills a quarter of the quantization register, so this is repeated four times. \circled{2} Each time the quantization register becomes full, it is written back to the row with $Q(\theta)$. It is repeated for the consecutive columns of $\theta$.

% \subsubsection{All-Reduce}
% For distributed data parallel training, 
% the minibatch is split into multiple learners (a chip in this case) and the gradients are averaged over all learners.
% A popular implementation of all-reduce is the ring algorithm~\cite{tharkur} and in each learner's perspective, 
% each NPU receives a portion of the gradients, and performs a parallel addition on the portion with the local copy of the gradients.
% This can be easily executed in-memory by \pimname and we omit the details for the space limitations.
% \vspace{-5mm}
\subsection{Commanding \pimname}
\label{sec:command}
\JLr{because of deterministic latency}
We utilize the RFU (Reserved for Future Use) commands in existing DDR4 protocol~\cite{DDR4} to realize the \pimname commands without modifying the existing commands. According to the standard, there are a number of configurable command signals for RFU operations. Since all the commands require addresses for bank groups, banks, rows, and columns, it leaves five signals left for configuring \pimname commands
%\footnote{There are four RFUs commands in total and signals CA5, CA9-12 are left unused. Also. Please refer to \cite{ddr5} for more details}. 
\footnote{\vspace{-3mm} These are A12/BC\_n, A17, A13, A11 and A10/AP~\cite{DDR4}.} 
%In the case of DDR5, the effective number of free bits increase to six as there are four RFU commands with four free bits each~\cite{ddr5}. 

\tablename~\ref{tbl:truth} shows the truth table for the commands added for \pimname.
For scaled read, we assign 2 bits for the scaler value id and 1 bit for the dst register id.
For quantization and dequantization, 2 bits are assigned to choose the offset within the quantize register (to support up to four times quantization ratio), and another bit for the src/dst temporary register id.
For writeback and parallel ops, 1 bit is assigned to denote the src or dst register id. 
Parallel ops do not require src register ids since there are only two temporary registers that are both used as operands.
With quantization register control, we assign a single bit for wr/rd.

\begin{table}[bp] 
\small
%\footnotesize
	\centering
	\vspace{-4mm}
	\caption{Truth Table for \pimname Commands}
	\begin{tabular}{|@{\hspace{0.2em}}l@{\hspace{0.2em}}|c|c|c|c|c|}
		\hline
		\hline
       \backslashbox{\footnotesize{Func.}}{\hspace{-0.2em}\footnotesize{Signal}} & Op0 & Op1 & Param0 & Param1 & Src/Dst \\

		%\midrule
		\hline
		\hline
        Scaled Read & L & L & \multicolumn{2}{c|}{Scale ID} & Dst \\
        DeQuant     & H & L & \multicolumn{2}{c|}{Src Position} & Dst \\
        Quant       & H & H & \multicolumn{2}{c|}{Dst Position} & Src \\
        Writeback   & L & H & L & L & Src \\
        
        Q. Reg      & L & H & H & L & RD/WR \\
        Add         & L & H & H & H & Dst \\
        Sub         & L & H & L & H & Dst \\
		\hline	
		\hline	
	\end{tabular}
	%\vspace{-6mm}
	\label{tbl:truth}
\end{table}

%Among those we assign four bits for opcode to handle scaled read, register store, parallel arithmetic ops and writeback (including with scaler value ids and the arithmetic operation type) and 1 bit for the target register. 
%We do not assign operand register ids, because we always use both temporary registers as the operands. In case of the writeback command, the target register id is used as the read-from register.

%s, one bit for each of the two sources and the destinations.\JL{dont we need three registers???? src1, src2 and dst?? - this gives 2 registers in total and contradicts with the fixed scaling I have explained in 3.1} 
In case we need more fields to support additional operations for future extensions, we can add an extra command signal or occupy unused command combinations which are not explicitly stated as RFUs, but not claimed by the standard. This would cause slightly more overhead to complicate the command decoder design, but would provide a plenty of command signals for enough flexibility.

\begin{figure}
\centering
 \includegraphics[width=.85\columnwidth]{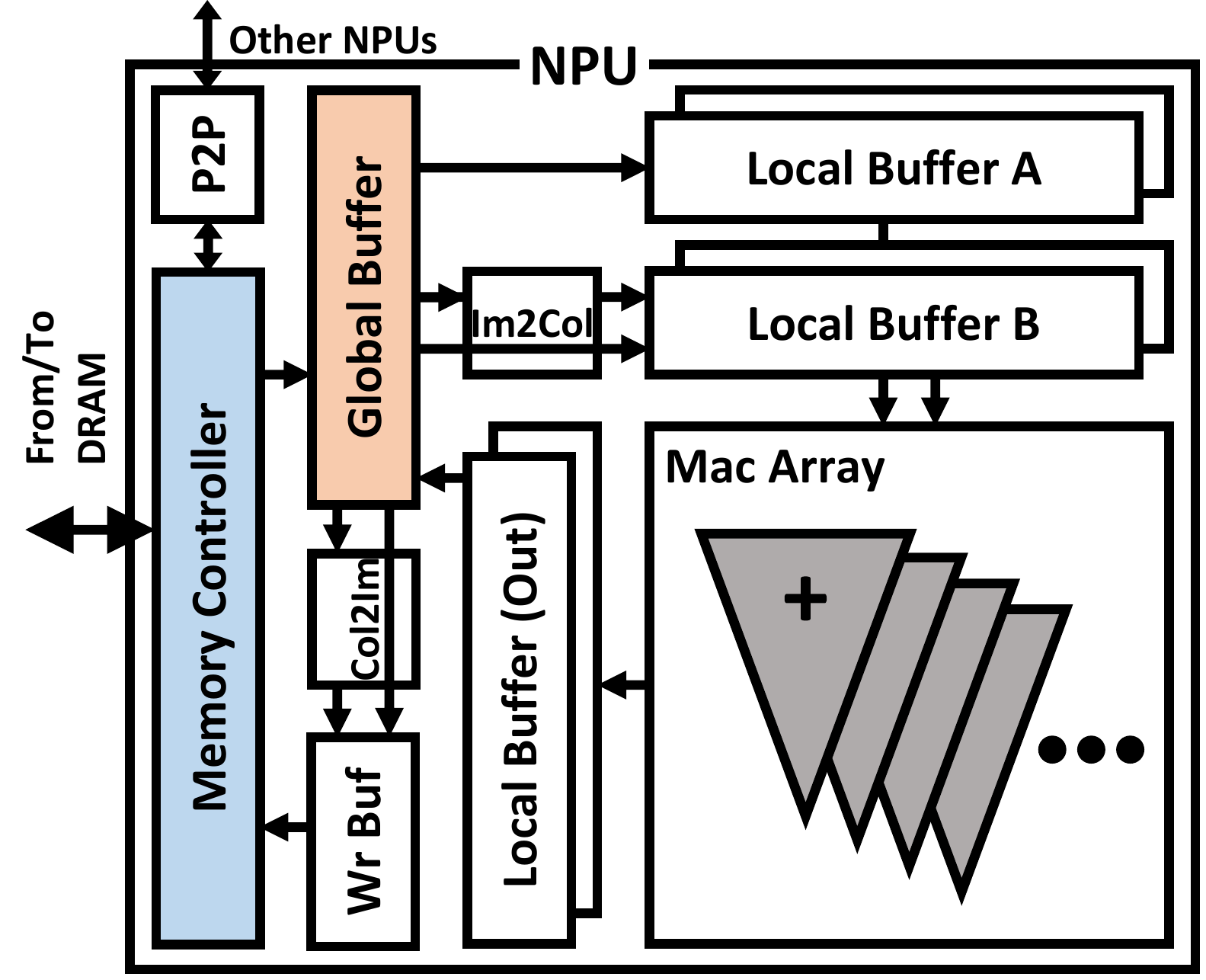}
 \caption{NPU architecture used with \pimname.}
 \vspace{-3mm}
 \label{fig:npu}
  \end{figure}

\section{System Design for \pimname}
\subsection{NPU Architecture}

We design a Diannao-like NPU as shown in \figurename~\ref{fig:npu}.
Modern NPUs often utilize systolic array structures for MACs. However, employing systolic arrays make it necessary for the NPU to perform additions in a sequential way. When using low-precision values, this often leads to numerical stability problems due to swamping~\cite{swamping}.
A popular way of solving this issue for low-precision training is chunk-based additions~\cite{ibm1}, which is a method that gradually adds up the elements in chunks so that there is less divergence between the exponents of the partial sums.
For this reason, we compose the \emph{MAC} array of our proposed NPU as 256 multiplier-adder trees where each tree receives 256 pairs of input values and calculates the sum of products to output one activation.

To feed the MAC array with continuous streams of data, we adopt the widely used im2col-col2im dataflow that converts the input activations into a Toeplitz matrix so that convolution operations can be converted into matrix-matrix or matrix vector multiplications as done by cuDNN and a few NPUs~\cite{mbs, cudnn}.
The input matrices are partitioned into 256x256 blocks that fit the local buffer in order to maximize data reuse.

There are two sets of input local buffers (one for weight parameters and the other for activations in forward pass) and a set of output local buffers per adder tree.
Each input buffer holds the 256x256 elements of a block, which is equal to the size of inputs of the MAC array. The output buffer is also 256x256, matching the size of the partial sum of the resulting matrix block.
Each input and output local buffer are double-buffered to maintain throughput while reading/writing from/to the global buffer.
Each cycle, the local buffer of weight parameters provides a row of blocked matrix to each tree, and the activation local buffer provides a column of the blocked matrix to each adder tree. 
At the end of each cycle, the columns in the local buffer rotate, making a different match between the rows and the columns to be multiplied. 
This is analogous to the weight-stationary dataflow~\cite{ws1, ws2} often used in systolic array architectures~\cite{shidiannao, tpu, eyeriss}.
When the partial sum for the entire block is complete, the value is written back to the global buffer, while processing of the next block proceeds using the values prepared in the other side of the double-buffers.

In order to avoid the memory traffic explosion due to the im2col scheme, we place a dedicated module for performing im2col. % between the global buffer and the activation local buffer. 
The module creates a block of matrix from the image-format activations in the global buffer and writes to the activation local buffer.
The global buffer aggregates a few matrix blocks to form a macroblock so that it can hold the right amount of data at a time.
The calculated output blocks coming from the MAC arrays are accumulated in the global buffer. When processing is done for the macroblock, it is sent to the DRAM through the write buffer. In case of the backward phase, a dedicated col2im module is used between the global buffer and the write buffer, performing the inverse of an im2col required for the backward pass.\JLr{@Heesu, plz thoroughly check} 

\JLr{adder between the local buffer and the global buffer?}

\begin{figure}
 \centering
 \includegraphics[width=1\columnwidth]{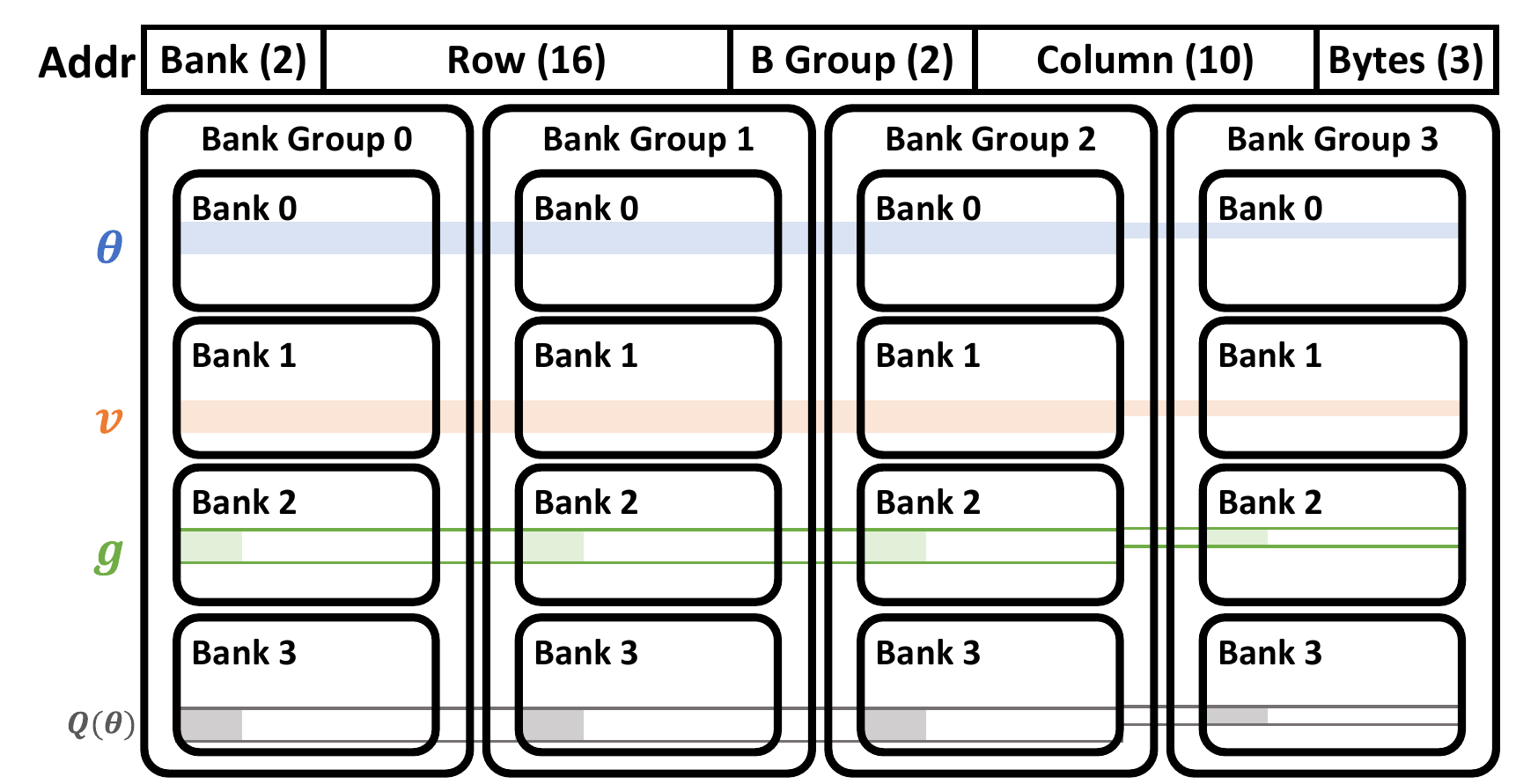}
 \caption{Address mapping and data placement scheme.}
 \vspace{-2mm}
 \label{fig:mapping}
  \end{figure}

\subsection{Data Placement}
\label{sec:placement}
Conventional DRAM subsystems integrate multiple devices to form a data bus with a large data width. 
64-bit data are split into 4, 8 or 16 bits and interleaved among devices for a x4, x8, and x16 device respectively.
However, \pimname requires the entire 32bit within a device in order to perform vector operations.
We simply follow the non-interleaving data arrangement scheme used in~\cite{nda, bufcmp} where consecutive bits are placed within each device. This puts an entire word into a device and allows for element-wise operations.

Considerations are needed for the address mapping to avoid bank-group and bank conflicts.
Except for the simple SGD, update phase requires more than one value per parameter (e.g., $\theta$ and $v$ in Eq~\ref{eq:momentum_theta}). 
Reading them are sequential, but incurs a cumbersome problem for \pimname. 
When the two arrays of values are placed in two different bank groups, it requires an inter-bank communication, which %is unsupported by \pimname because it 
occupies the global shared data bus. %within the DRAM chips. 
On the other hand, if the two arrays are placed within a single bank, it is a trivial case of a bank conflict. %It would require alternating row activations for accessing each array.

Therefore, it is necessary to ensure that the corresponding elements of the arrays are placed within the same bank group, but to different banks. It can be solved by carefully designing the address mapping. \figurename~\ref{fig:mapping} shows the address mapping for \pimname. To enable maximum bank-group level parallelism, we adopt bank-group interleaving, so that multiple bank groups can operate concurrently. 
The bank ids within the bank groups are assigned to the MSB of the addresses.
This makes sure that multiple different arrays can always be placed in distinct banks. 
When allocating the arrays such as $\theta$ or $v$, they are aligned to the bank boundary, so that the items at the matching positions always stay within the same bank group.

For the quantized weight parameters, it is impossible to perfectly align them with the non-quantized weight parameters, because their elements differ in width. % while the number of elements are equal to the other values. 
If they are aligned to the beginning of the array, they will not be present in the same bank group anymore. %, violating the \pimname operation requirements.
To solve the problem, we choose to utilize only the first quarter (for 8/32 bit quantization) of the row for the quantized weights. 
By doing this, even though we might waste the DRAM capacity, we do not waste the off-chip bandwidth. 
The \pimname operation will run without a problem because the column id of the elements do not have to match as long as they are placed within the same bank group and different bank.
We do not consider multiple channels/ranks in this section, but the channel or rank bits can be placed between the bank group bits and the bank bits as long as it does not violate the same bank group, different bank criteria.

\begin{table}[bb!]
\small
\centering
	\vspace{-4mm}
\caption{DRAM parameters }
\begin{tabular}{lclc}
\toprule
\multicolumn{2}{c}{Spec.} & \multicolumn{2}{c}{DDR4-2133} \\
%\multicolumn{2}{c}{Process} & \multicolumn{2}{c}{22nm}\\
\midrule
%\multicolumn{2}{c}{Timing} & \multicolumn{2}{c}{Energy} \\
   Timing (Cycles)	& Value & Current (mA)			 & Value \\
\midrule
tCK	&	0.94ns	&	Vdd	&	1.2V	\\
tCL	&	16	&	IDD0	&	75	\\
tRCD	&	16	&	IDD2P	&	25	\\
tRP	&	16	&	IDD2N	&	33	\\
tRAS	&	36	&	IDD3P	&	39	\\
tCCD\_L	&	6	&	IDD3N	&	44	\\
tCCD\_S	&	4	&	IDD4W	&	225	\\
tRR	&	1	&	IDD4R	&	225	\\
tPIM	&	5	&	IDDpre	&	98	\\
%tFAW 23 
 \bottomrule
\end{tabular}
\label{tbl:parameters}
\end{table} 

To exploit the effect of data mapping to the programming model, 
we assume a device-side allocation function supporting separation between data structures is provided, similar to multi-stream features in SSDs~\cite{multistream}. 
Therefore the user only specifies that each data structure should be stored to different banks, without being directly exposed to the the physical banks or bank groups.
In the current setting, we assume that the NPU has its dedicated memory attached with \pimname. However, \pimname can also be used when the host and the NPU share the memory by assigning certain memory region to the NPU, similar to pinned memory in CUDA~\cite{cuda}.

\rindex{R2a}
\rev{An alternative to the aforementioned data placement 
is to use array-of-structures, where multiple types of parameters form a structure that are repeatedly and sequentially stored in the memory as arrays. 
While this solves many of the problems such as bank conflict or inter-bank communication for the parameter update phase, it has a critical drawback for the forward and backward phase for having to read unnecessary parts of data within the structure along with the desired ones. %, because data is read from DRAM in 64B bursts. 
We demonstrate its effect in Section~\ref{sec:exp}.
}

\begin{figure}
 \includegraphics[width=\columnwidth]{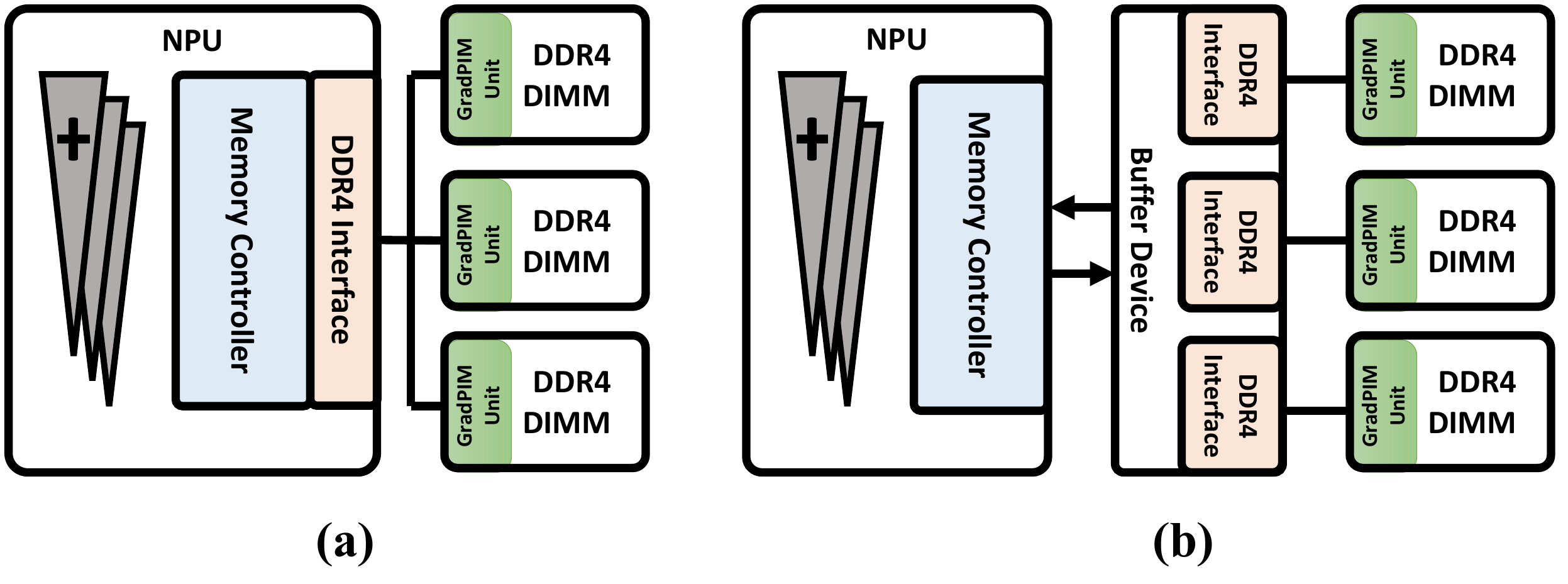}
 \caption{Interfacing NPU with the memory system. (a) \pimname-Direct. (b) \pimname-buffered.}
 \vspace{-3mm}
 \label{fig:interface}
  \end{figure}
  
\begin{figure*}
    \centering
  %\includegraphics[width=0.6\textwidth]{figs/legend_simtime.pdf}
 %\vspace{-8mm}
 \includegraphics[width=\textwidth]{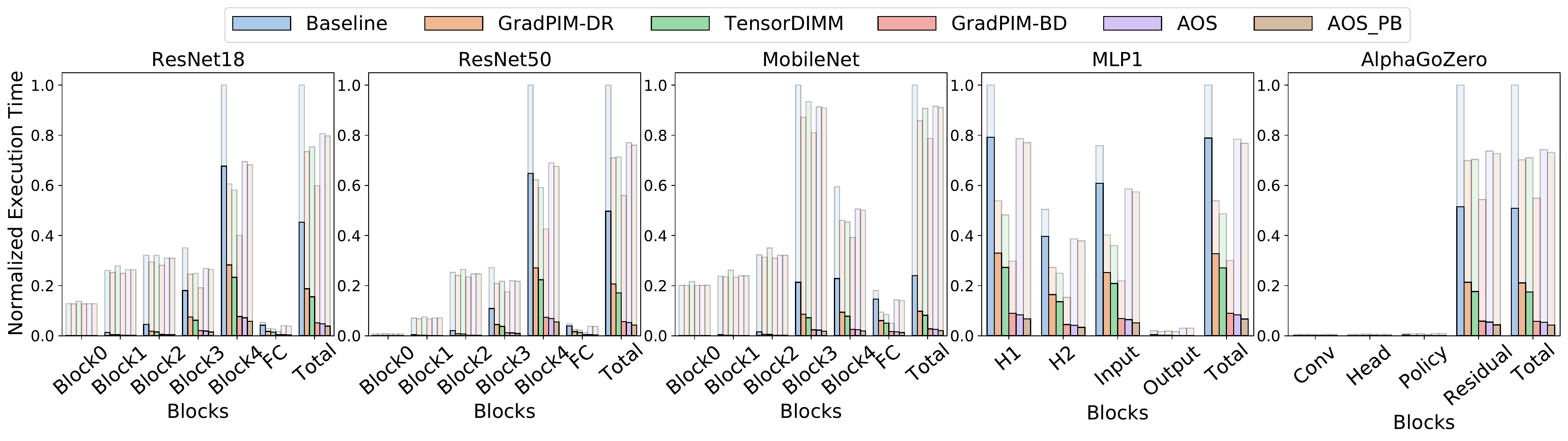}

 %\vspace{-2mm}
 \caption{\rev{Normalized execution time of each layer on various networks using \pimname. The filled parts of the bars represent parameter update phase, and the translucent parts of the bars represent the forward/backward phase.} }
 \label{fig:perf}
 %\vspace{-5mm}
  \end{figure*}

\subsection{Memory System Interfacing}
\label{sec:interface}
Typically, a DDR4 memory chip communicate with its host chip through memory controllers that follows the DDR4 protocol, which are placed on the host chip. 
Such systems are called direct-attach memory system, and is shown in \figurename~\ref{fig:interface} (a) (\emph{\pimname-Direct}).
While it is a popular design choice, our experiments show that the command bus becomes a bottleneck that prevents \pimname from exploiting the full bank-group level internal bandwidth from the DRAM chips (See \figurename~\ref{fig:subperf}).

In light of the fact, we design an alternative version of \pimname based on buffered memory system as shown in \figurename~\ref{fig:interface} (b) (\emph{\pimname-Buffered}).
Often used for high-end servers, buffer devices are placed between the host chip and the memory devices. 
\rindex{R8}\rev{They usually function as fan-out expander by buffering and repeating signals.}
%While buffer devices and the DRAM devices communicate using the memory protocol (i.e., DDR4), 
The host chip and the buffer device communicate via serial interconnects such as OMI~\cite{opencapi} or AMB interface~\cite{fbdimm}.
Buffer devices have separate command buses that can mitigate the command bus bottleneck.
High-level commands are defined such as those in HMC~\cite{hmc} in order to reduce the host-to-buffer device command bandwidth.

\subsection{Distributed Learning}

Distributed data parallelism~\cite{horovod, ddp} allows using multiple nodes to run the exact same model with different parts of the minibatch.
This provides another opportunity for \pimname. 
By applying distributed data parallelism, it parallelizes the forward and backward pass of the training, but the parameter update phase is performed independently at each NPU, almost equivalent to the sequential portion of the application. 

To support distributed training with multiple NPUs, a dedicated P2P unit is placed, so that it can perform communications to other NPUs via high-speed interconnects similar to GPUDirect~\cite{gpudirect}. As in many GPU-based distributed learning work~\cite{224sec, 4min}, we use all-reduce~\cite{tharkur} communication pattern. 
The all-reduce communication includes accumulating the gradient in the every gradient sharing step, which is also mapped to \pimname similar to the update procedures.

\begin{table}[bp] 
\small
%\footnotesize
	\centering
	\vspace{-4mm}
	\caption{Layout Results}
	\begin{tabular}{|@{\hspace{0.2em}}l@{\hspace{0.2em}}|c|c|c|}
		\hline
		\hline

		 Module & Area ($\mu m^2)$ &	Power (mW) \\
		%\midrule
		\hline
		\hline
		%\quad PSL & 82350 && 76480 &\\
	   Adder & 320.1 & 0.058 \\
	   Quantize & 275.4 & 0.056 \\
	   Dequantize & 244.8 & 0.041 \\
	   Scaler & 606.1 & 0.159 \\ 
	   Registers ($\times$3) & 206.7 & 0.04 \\

       \hline
		%\hline
		Total & 8267.8 & 1.74 \\ 
		\hline	
		\hline	
	\end{tabular}

	\label{tbl:synth}
\end{table}

\section{Evaluation}
\label{sec:eval}
\subsection{Evaluation Methodology}
We evaluate the proposed NPU with an in-house simulator written in SystemC~\cite{systemC}. 
\pimname is modeled by extending DRAMSim 3.0~\cite{dramsim3}, and we faithfully modeled the timing of \pimname operations as explained in Section~\ref{sec:timing} while keeping the existing timings posed by the existing DDR protocol. 
To verify the NPU, we synthesized the NPU with 8bit datapath and 256$\times$256 MAC-adder trees in Verilog HDL at 1GHz using Nangate 45nm open cell library~\cite{nangate}.
We have ensured the timing closure and the functionality % then drew the power consumption 
of the NPU. % from a gate-level simulation.
 
Unless noted otherwise, DRAM devices are based on DDR4-2133 with 4 ranks, 4 bank groups and 4 banks per bank group.
 We set the bandwidth of the serial interconnect for the buffer devices (when used) to be equal to that of the \pimname-Direct for a fair comparison.  
We referred to the energy and timing parameters from \cite{micron}. The key parameters are displayed in \tablename~\ref{tbl:parameters}.
The overhead of the \pimname unit within the memory is modeled by conducting a layout at 45nm technology over the DRAM constraint of 3 metal layers and 70\% core utilization, which is then scaled to 32nm. 
The area and energy measured is shown in \tablename~\ref{tbl:synth}. 
When implemented in an x8 8Gb DDR4-SDRAM device, \pimname only incurs 0.01\% area overhead to the DRAM, which approximately corresponds to the size of a 1Mb DRAM cell.
We followed \cite{fgdram} to model the partial energy needed for read/write within the bank group (IDDpre). 
For the off-chip link model, we use the DRAM power calculator provided by Micron~\cite{microncalc}. We evaluate our design with two versions of ResNet (18- and 50- layers), Mobilenet~\cite{mobilenet}, MLP~\cite{mlp} and AlphaGo Zero~\cite{alphagozero} to represent various types of DNN workloads.
%The results are shown in \figurename~\ref{fig:perf} - \ref{fig:subperf}.
 %For the baseline, we have used the proposed NPU, with 8 bit datapaths with dedicated 32 bit update units within the NPUs. 

\subsection{Experimental Results}
\label{sec:exp}

 \textbf{Performance.} 
 \figurename~\ref{fig:perf} shows the execution time of the chosen networks for minibatch size 32 (128 for MLP). The layers are grouped into a few blocks with similar SRAM requirement characteristics for brevity.
 In the figure, the filled parts of the bar represent parameter update phase, and the translucent parts of the bars represent the Fwd/Bwd phase.
The leftmost bar shows the baseline, where the NPU has dedicated 32bit modules to execute the update phase including adders and quantize/dequantize units to convert between the precisions. 

%The times are separated into `Pup', `Fwd', `Bact', and `Bwgt' that represent parameter update, forward, backward activation, and backward weights respectively.

`\pimname-Direct' and `\pimname-Buffered' represent the two ways of interfacing the memory system as described in Section~\ref{sec:interface}. 
`TensorDIMM' represents a design similar to TensorDIMM~\cite{tensordimm} (also close to \cite{practicalpim, mcn})  where buffer chips are added between the NPU and the memory, and parameter updates are executed within the buffer chips.  
\rindex{R2b}\rev{Also, `AoS' represents a naive PIM design, where array-of-structures placement explained in Section~\ref{sec:placement} was used on top of `\pimname-Buffered' configuration. `AoS-PB' uses the same design, but places a \pimname unit \emph{Per Bank}, at the expense of more area and design burden.}
The execution times are normalized to the baseline execution time of the most time-consuming block within each network, while the `total' is separately normalized to the baseline execution time of the entire network.

Compared to the baseline, \pimname-Direct achieves around \num{2.25}$\times$ higher performance for the parameter update phase from utilizing the internal bandwidth of DRAM bank groups. 
For the entire training, the overall speedup is about \num{1.38$\times$} in geometric mean as it is governed by the Amdahl's law.
With \pimname-Direct, the bottleneck is the limited command rate (Section~\ref{sec:command}).
TensorDIMM and \pimname-Buffered alleviate the problem by using buffered DIMMs. % and sending the commands from the buffer devices.
TensorDIMM achieves \num{1.36$\times$} speedup in geometric mean. 
However, it's speedup is limited by the amount of rank-level parallelism.
\pimname-Buffered achieves \num{1.94$\times$} speedup overall, and \num{8.23$\times$} on parameter update by utilizing the bank group level parallelism.

\begin{figure}
\centering
\includegraphics[width=.95\columnwidth]{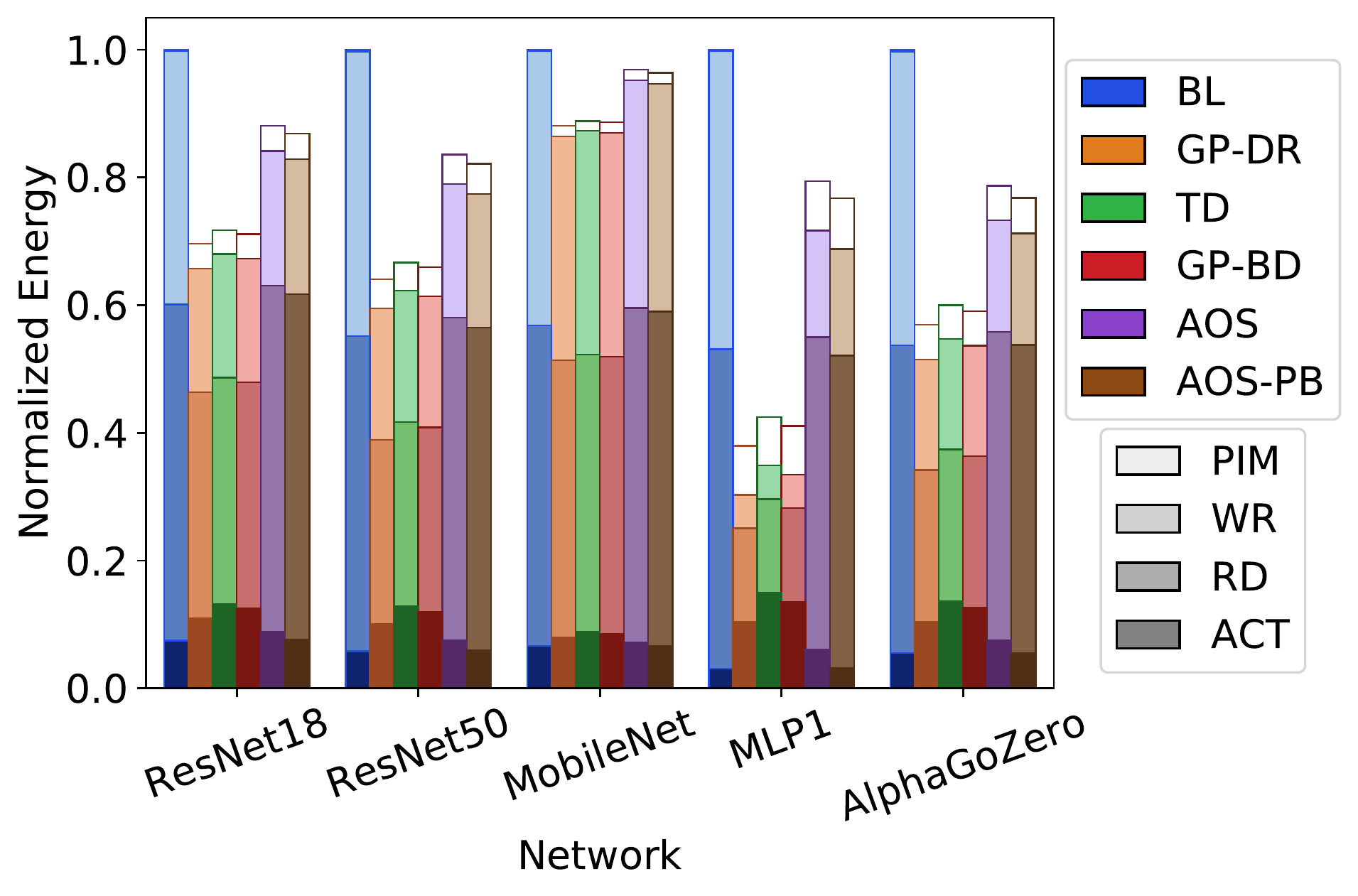}
\vspace{-1mm}
\caption{\rev{Energy consumption of \pimname and other techniques.}}
\vspace{-3mm}
\label{fig:energy}
\end{figure}

%\subsection{Alternative Designs}
\rindex{R2c}\rev{In \figurename~\ref{fig:perf}, AoS achieves similar speedup on parameter updates, having the same internal bandwidth as the \pimname-Buffered. However, it reduces the effective bandwidth of Fwd/Bwd to 1/4, because unnecessary to-be-discarded data will be mixed inside every DRAM burst. 
%The slowdown is lower than 4$\times$ because Fwd/Bwd is relatively a compute-intensive phase, but 
As a result, most of the benefit from using \pimname is diminished.
}

\rindex{R3}\rev{
AoS-PB shows the performance when a \pimname unit is placed per bank.
Putting aside the increased area overhead, the per-bank design can increase the speedup on the parameter update with high bank-level internal bandwidth ($4\times$ in DDR-4). 
%However, it is forced to use AoS placement as only one row can be activated at a time, and suffers from the slow Fwd/Bwd as in the AoS case. 
However, since only one row can be activated at a time, AoS placement is mandatory, from which the model will suffer from the same burst inefficiency.
As a result, the $4\times$ inefficiency in the Fwd/Bwd dwarfs the small benefit of the per bank design. % in the parameter update phase.
}

%For all the networks, the gain is larger as it goes towards the later layers since the ratio of the number of parameters to the amount of computation becomes higher as feature dimensions decrease and channel depths increase.
%\rev{@@analysis on other networks and the overall speedup is about 13\%.}

%, trim=0 -6cm 0 0]{figs/legend_energy.pdf}

%
\begin{figure*}
    \centering
    \begin{minipage}{0.33\textwidth}
        \centering
 \includegraphics[width=1.0\textwidth]{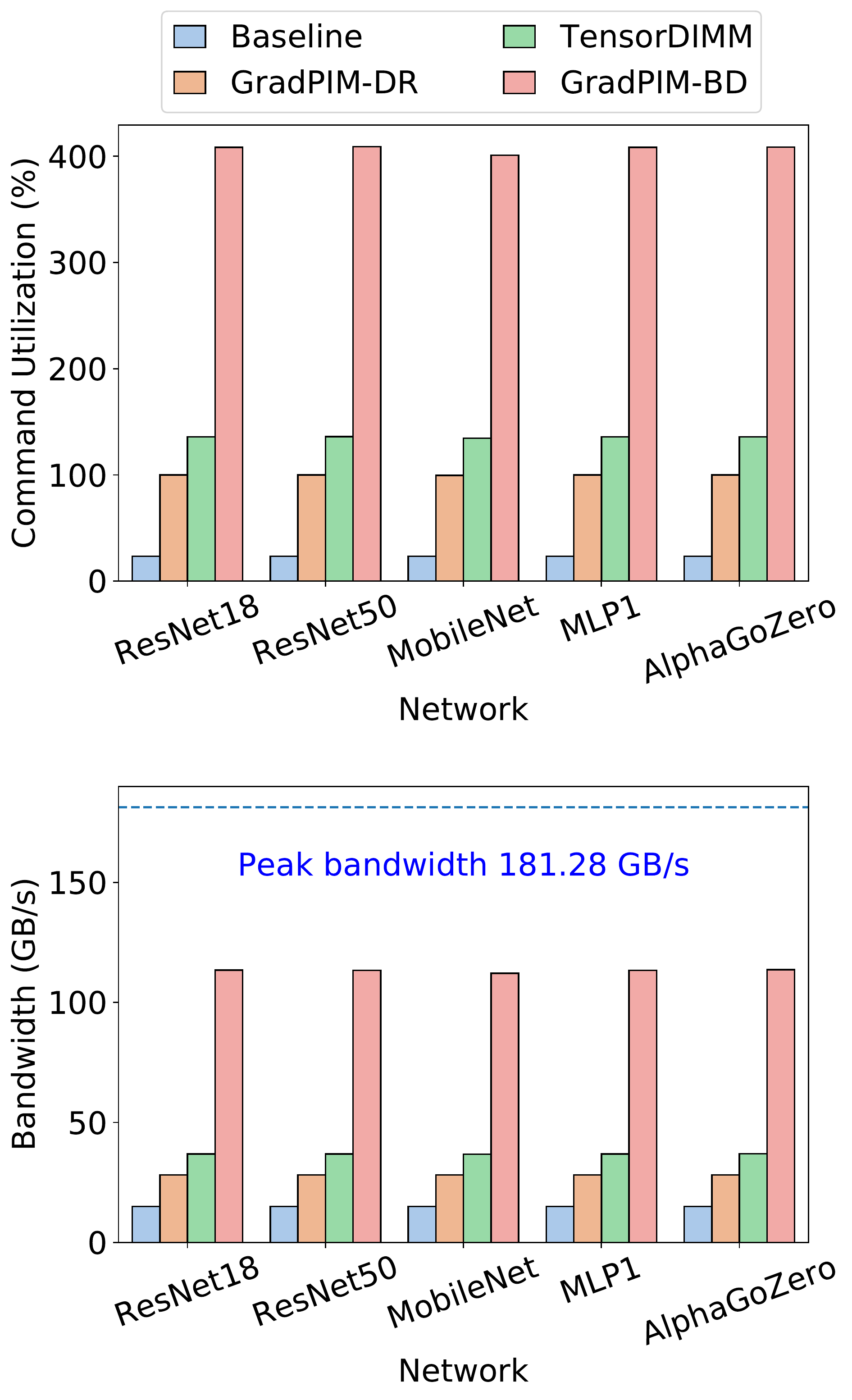}

% \hspace{1cm}
% \includegraphics[width=0.8\columnwidth]{figs/legend_simtime.pdf} 
% \includegraphics[width=\columnwidth]{figs/bottleneck.pdf} 
 \caption{Command bus utilization (top) and the \rev{internal memory bandwidth consumption using \pimname (bottom).}}
 \label{fig:subperf}
    \end{minipage}
    \hfill
    \begin{minipage}{0.65\textwidth}
        \centering

\subcaptionbox{\label{fig:sense:ratio}}{\includegraphics[width=0.55\textwidth]{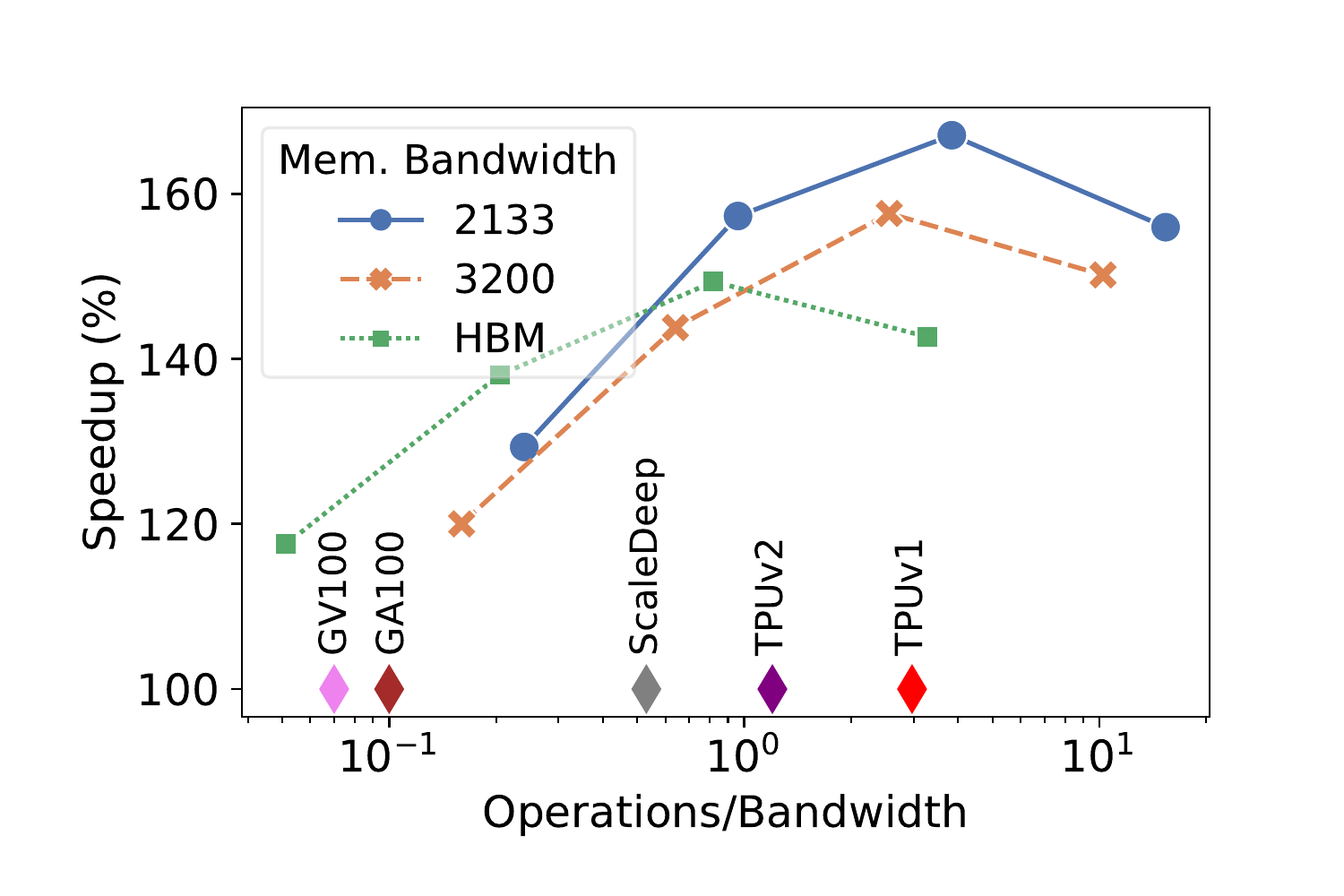}\hspace{-9mm}}
\subcaptionbox{\label{fig:sense:batch}}{\includegraphics[width=0.55\textwidth]{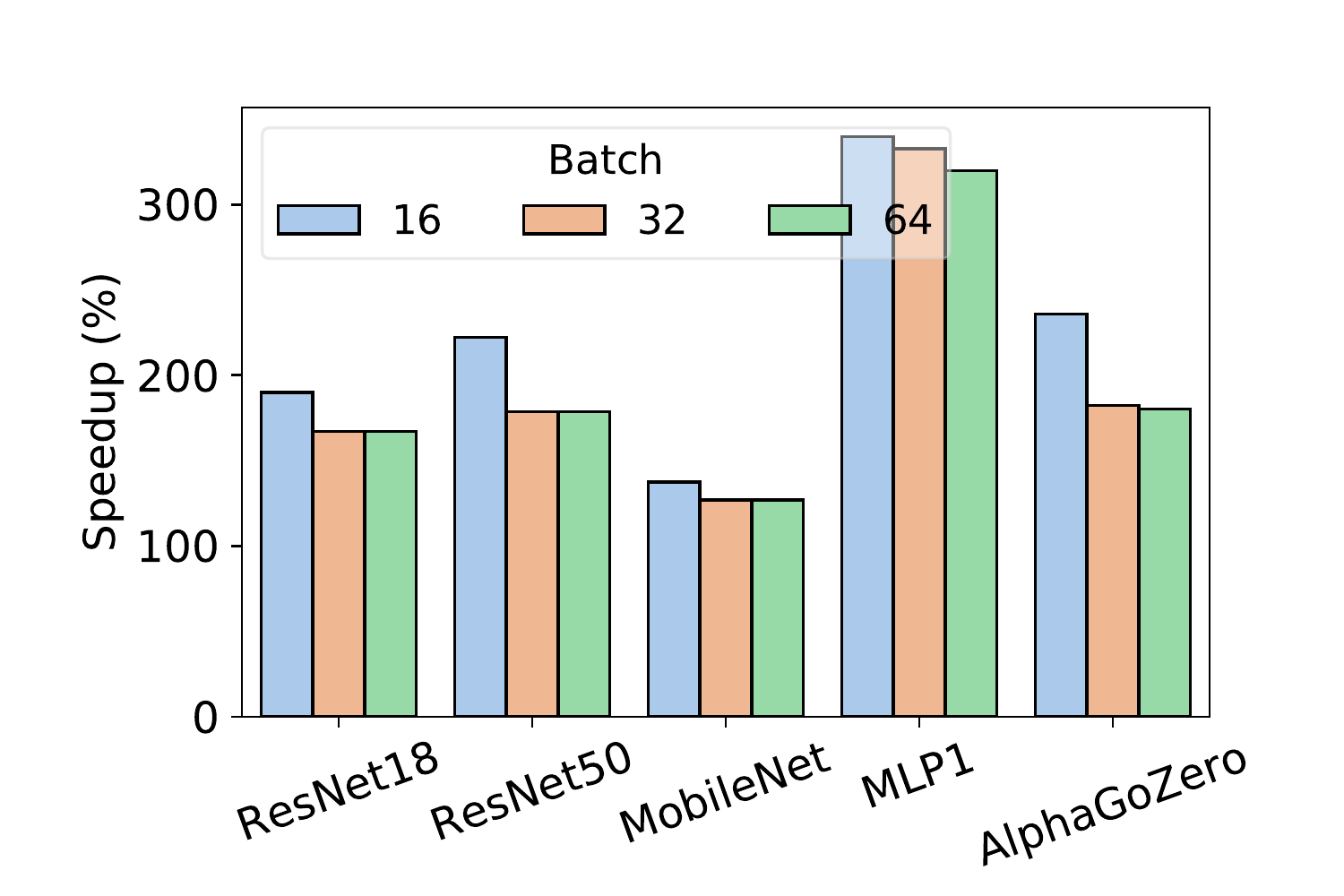}\hspace{-5mm}}
%\newline
\subcaptionbox{\label{fig:sense:prec}}{\includegraphics[width=0.55\textwidth]{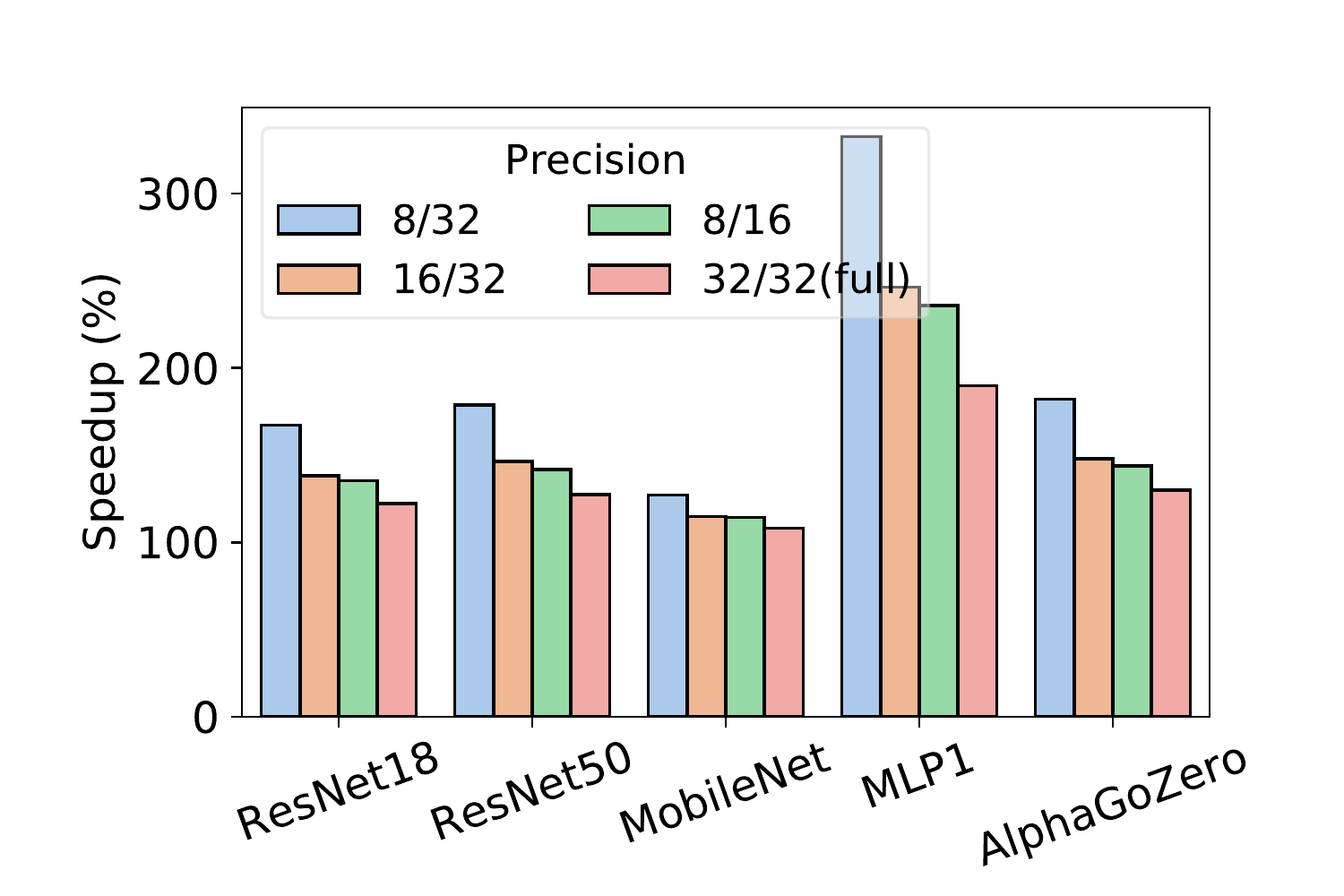}\hspace{-9mm}}
\subcaptionbox{\label{fig:sense:precE}}{\includegraphics[width=0.55\textwidth]{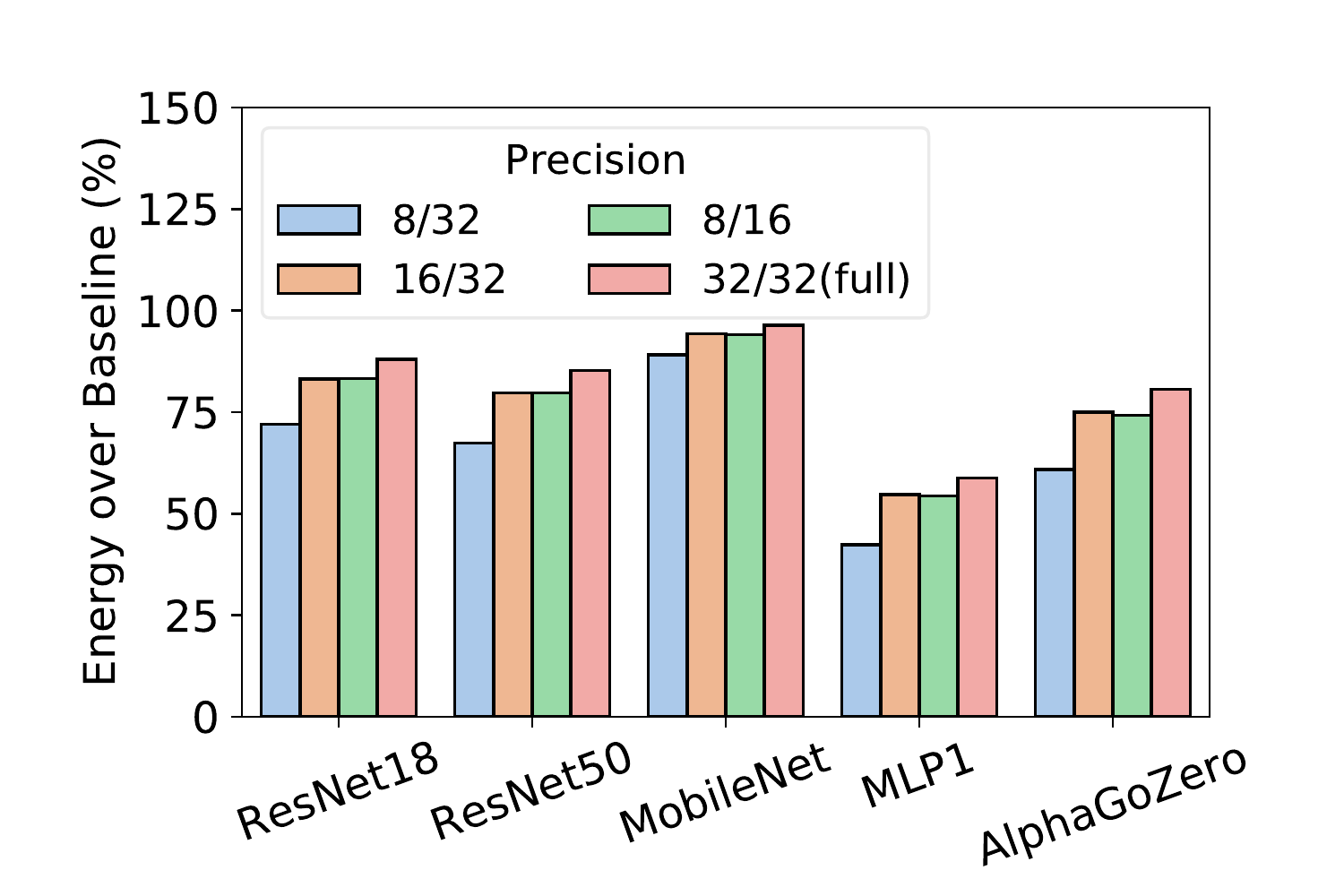}\hspace{-5mm}}
\hspace{55mm}

 \caption{Sensitivity to compute-bandwidth ratio (a), minibatch size (b), and precision mix ((c)-speedup and \rev{(d)-energy)}.}
    \end{minipage}
\end{figure*}

% \begin{figure*}
%   \centering

% %\includegraphics[width=\textwidth]{figs/fig12_2.pdf}
% % \vspace{10mm}
% \subcaptionbox{\hspace{0mm}\label{fig:sense:ratio}}{\includegraphics[width=0.35\textwidth]{figs/0_sense2.pdf}}\hfill
% \subcaptionbox{\hspace{0mm}\label{fig:sense:batch}}{\includegraphics[width=0.35\textwidth]{figs/0_batch2.pdf}}\hfill
% \newline
% \subcaptionbox{\hspace{0mm}\label{fig:sense:prec}}{\includegraphics[width=0.35\textwidth]{figs/prec2.pdf}}\hfill
% \subcaptionbox{\hspace{0mm}\label{fig:sense:precE}}{\includegraphics[width=0.35\textwidth]{figs/precE2.pdf}}\hfill
% % \vspace{10mm}
% %  \subcaptionbox{\hspace{-10mm}\label{fig:sense:ratio}}{ \vspace{-2mm}\includegraphics[width=0.235\textwidth]{figs/0_sense.pdf}
% %  }\hfill%\hspace{-3mm}
% %  \subcaptionbox{\hspace{-5mm}\label{fig:sense:batch}}{ \includegraphics[width=0.23\textwidth]{figs/0_batch.pdf}
% %  }\hfill%\hspace{-3mm}
% %   \subcaptionbox{\hspace{-7mm}\label{fig:sense:prec}}{ \includegraphics[width=0.23\textwidth]{figs/prec.pdf}
% %  }\hfill%\hspace{-3mm}
% %   \subcaptionbox{\hspace{-7mm}\label{fig:sense:precE}}{ \vspace{-.37mm}\includegraphics[width=0.238\textwidth]{figs/precE.pdf}
% %  }\hfill%

%  \caption{Sensitivity to compute-bandwidth ratio (a), minibatch size (b), and precision mix ((c)-speedup and \rev{(d)-energy)}.}
% %\vspace{-2mm}
 
%  \label{fig:sense}
%   \end{figure*}

   \textbf{Energy Consumption.} 
The energy consumption from the memory is shown in \figurename~\ref{fig:energy}. 
\rindex{R4a}\rev{Each design is depicted with different colors, while the breakdown between PIM, Write, Read and Activate are distinguished with the brightness, in the respective order. For TensorDIMM (TD), we show the energy consumption of a PIM unit on a buffer device, assuming a 32nm logic process buffer device.}
%In general, the energy consumed from the NPU remains similar as most of the energy is consumed from the MAC arrays, which does not change much from the baseline. 
%Most of the energy difference comes from the static power due to the decreased overall execution time on NPU with \pimname.
The energy saving is almost proportional to the speedup as the saving mostly comes from reduced external bandwidth usage, which translates to less switching. 
As illustrated in the breakdown, most of the energy reduction comes from the reduced amount of read/write.
%However, much energy saving comes from the memory, especially on the off-chip data bus.
For the parameter update phase of the non-PIM approach, the NPU reads high precision master weight, performs update, and writes the master weight back to memory, which all incorporates the off-chip bus.
On the contrary, \pimname only requires a one-way transaction of low-precision gradients from the NPU to memory.
Therefore, the proposed model benefits from two sources of savings: lower bit-width and reduced memory transaction. 

\rindex{R4b}\rev{The energy breakdown show that energy consumption of row activation is almost the same across all architectures, since \pimname does not change the number of data to be read from DRAM.
On the other hand, the energy consumption of RD/WR exhibit more variation, especially for AoS and AoS-PB where the increased data access during the Fwd/Bwd phase leads to much higher RD/WR count than \pimname-Direct and \pimname-Buffered.}
%In fact, the memory footprint is exactly the same, and the slight differences comes from change in the access order. 

%On the other hand, the DRAM internal energy consumption increases by almost @fold. 
%Since each bank group can load columns into registers without having to compete over the data bus, the bank group's internal bandwidth consumption sharply increases and leads to the energy consumption increase. 

%Taking all of these into account, the overall energy savings for the entire training with \pimname is @@20.5\% in the geometric mean. %\JLr{geometric mean?}.

   \textbf{Bottleneck Analysis.} 
\figurename~\ref{fig:subperf} (bottom) shows the internal bandwidth consumption of DRAM during the update phase. \rindex{R5a}\rev{It shows the bandwidth consumed inside the DRAM dies, and tells us the utilization of the \pimname units when compared to the peak internal bandwidth (dotted line)}.
The baseline NPU's external bandwidth consumption during the update phase is around 15GBps, reaching near the theoretical maximum of 17.1GBps.  
The internal bandwidth of the baseline is trivially the same.
On the other hand, 
the internal bandwidth consumption of \pimname-Direct on average is \num{28GBps}, far higher than that of the baseline.
The benefit comes from each \pimname unit working independently on each bank group, leading to speedup and decreased energy consumption.
\rindex{R5b}\rev{However, compared to the theoretical maximum of 181.28 GBps, it represents only 15.4\% of the peak bandwidth and the \pimname unit utilization.} %max 136.5
%However, the external bandwidth usage of \pimname is far lower than the maximum, implying that it is not the bottleneck.
%On the other hand, the internal bandwidth is not the maximum either, because we have in total 16 bank groups (from 4 ranks), providing 8$\times$ the external bandwidth which is 136.5GBps.
We found that the bottleneck is at the command bus, reaching near 100\% utilization as shown in \figurename~\ref{fig:subperf} (top). This means that the command bus is blocking any further internal bandwidth increase from \pimname-Direct.
%With one column read command, DDR4 protocol allows reading a 64B at a time. 
%With speed grade of 2133MT/s, the maximum internal bandwidth that can be achieved by the column read commands is around 70GBps, where our design is reaching at.
With \pimname-Buffered, the because commands are generated by the buffer chip, the command bus bottleneck is alleviated. 
It consumes around \num{113GBps}, almost \num{4.0}$\times$ the internal bandwidth compared to \pimname-Direct. 
%\figurename~\ref{fig:subperf} (left) shows that its command bus utilization reaches close to the maximum of the buffer device, implying that it is still the bottleneck. 
%We believe this can be mitigated using a self-refresh-like command generator placed per \pimname unit to reduce the command burden and increase the internal bandwidth utilization for further speedup.

\subsection{Sensitivity Analysis}
\textbf{Operations/Bandwidth ratio.} 
\rindex{R10}
\rev{To examine how \pimname would perform on accelerators in different environments,  
\figurename~\ref{fig:sense:ratio} shows the speedup sensitivity with respect to the operations/bandwidth ratio in a range that includes a few NPUs~\cite{tpu, tpuv2, scaledeep} and GPUs~\cite{gv100, ga100}.
The X axis shows the ratio between the operations and the memory bandwidth, using different MAC array sizes (64x64-512x512) and memory data rate (DDR4-2133, 3200 and HBM). 
The Y axis represents how much speedup is obtained with \pimname over baseline with each setting, measured from executing AlphaGoZero.}
% We have changed the tCCD\_L according to the values in \cite{micron}. 
%The X axis shows the ratio between the ops and the memory bandwidth. 
%We have marked the ratios of a few GPUs and NPUs~\cite{tpu, tpuv2, scaledeep, gv100, ga100}.
%For a fair comparison, we have used the metric of operations per second (compute) divided by activation item per second (memory bandwidth) to take different target precisions into account.
As Operations/Bandwidth of the NPU increases, \pimname achieves more speedup until the MAC array is too large that the latency to fill the array dominates the execution time, leading to diminished gains.
The figure shows that \pimname achieves meaningful speedups (20-70\%) for a range that covers the Operations/Bandwidth of the chosen NPUs, but the speedup diminishes as the ratio approaches that of the GPUs ($<$20\%).

\textbf{Minibatch size.} 
While the minibatch size does not affect the speedup  of \pimname over the update phase, it directly affects the portion of the update phase in the entire training. \figurename~\ref{fig:sense:batch} shows the change in the overall speedup coming from the batch size. 
While the overall speedup with regard to the minibatch size does not change by a large amount, it shows a continuous trend where smaller batch size leads to higher speedup.
Thus \pimname has better potential for speedup with smaller batch sizes. 
As we will see in Section~\ref{sec:ddl}, this will help improve the scaling efficiency for distributed deep learning.

\textbf{Sensitivity to mixed-precision levels.}
\rindex{R1b}\rev{\figurename~\ref{fig:sense:prec} and \ref{fig:sense:precE} shows the difference on speedup and energy consumption when 8/16 bit, 16/32 bit mixed-precision or full-precision (32/32 bit) is used instead of 8/32 bit as in our default setup.}
While the speedup of \pimname highly depends on the ratio between the low-precision and the high-precision representations, the setting of 8/16 bit, 16/32 bit, and 32/32 bit systems still provide meaningful speedups of \num{1.39$\times$}, \num{1.43$\times$}, and \num{1.26$\times$}, respectively.
\rindex{R1c}\rev{Also, \figurename~\ref{fig:sense:precE} shows the energy consumption, compared to the baseline no-PIM technique on each precision. 
It shows a similar trend, since most of the advantage on performance and energy both come from reducing the off-chip bus traffic. }
%While it shows a similar trend with the speedup, one thing to note is that sometimes \pimname applied on full precision shows a better energy consumption reduction ratio compared that of 8b/16b. 
%It comes from the fact that when in full precision training, the amount of data transfer that can be reduced by \pimname is larger, and results in a larger reduction in the off-chip bus. 

8-bit training is still at a premature status, and 16/32 bit mixed-precision training is dominant on the field. 
Despite this, we anticipate the onset of lower-bit training in the near future, where \pimname could bring many benefits to mixed-precision training.

  \begin{figure}
  \center
\vspace{-4mm}
 \includegraphics[width=0.85\columnwidth]{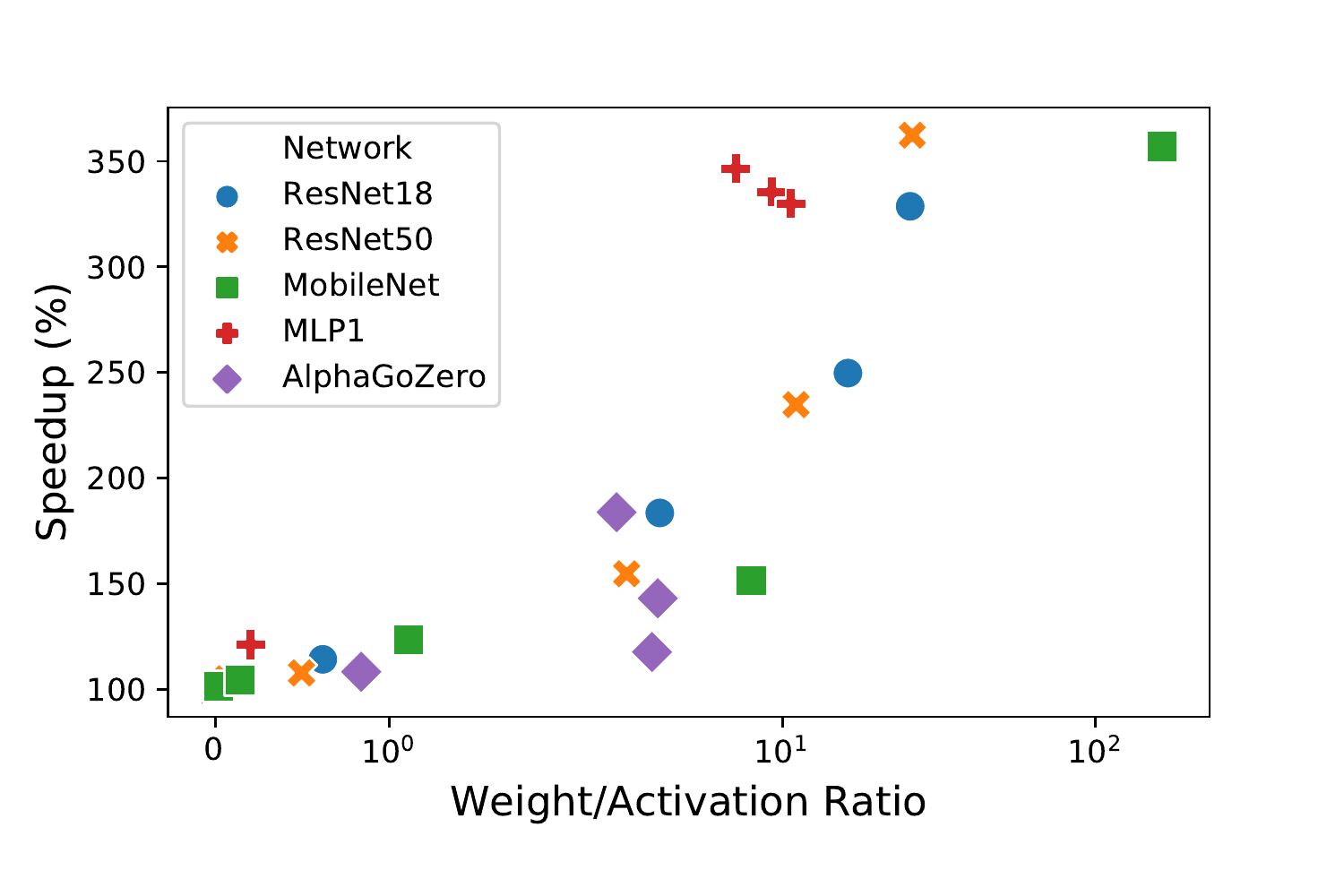}
\vspace{-4mm}
 \caption{Layer characterizations.\JL{fontsize, y axis number}}
% \vspace{-5mm}
 \label{fig:layer}
  \end{figure}

\subsection{Layer Characterizations.} 
To study the relation between the layer characterizations and the speedups obtained by \pimname, We show the speedups with regard to weight/activation ratio in \figurename~\ref{fig:layer} with the X axis in log scale. 
The plot shows a clear correlation between the weight/activation ratio and the speedup. 
Usually, for the convolutional layers from the earlier stages of the networks, there are large activation fields, with relatively small filters and the speedup for them are generally small.
For the layers from the later stages of the networks, the activation maps get smaller due to the result of repetitive pooling and strided convolutions. 
These convolutional layers and fully-connected layers often exhibit a very high weight/activation ratio, and the speedup gets larger.

  \begin{figure}
%  \includegraphics[width=\columnwidth]{figs/ddl_new.pdf}
    % \centering
% \vspace{-2mm}
 \hspace{2mm}
 \includegraphics[width=0.95\columnwidth]{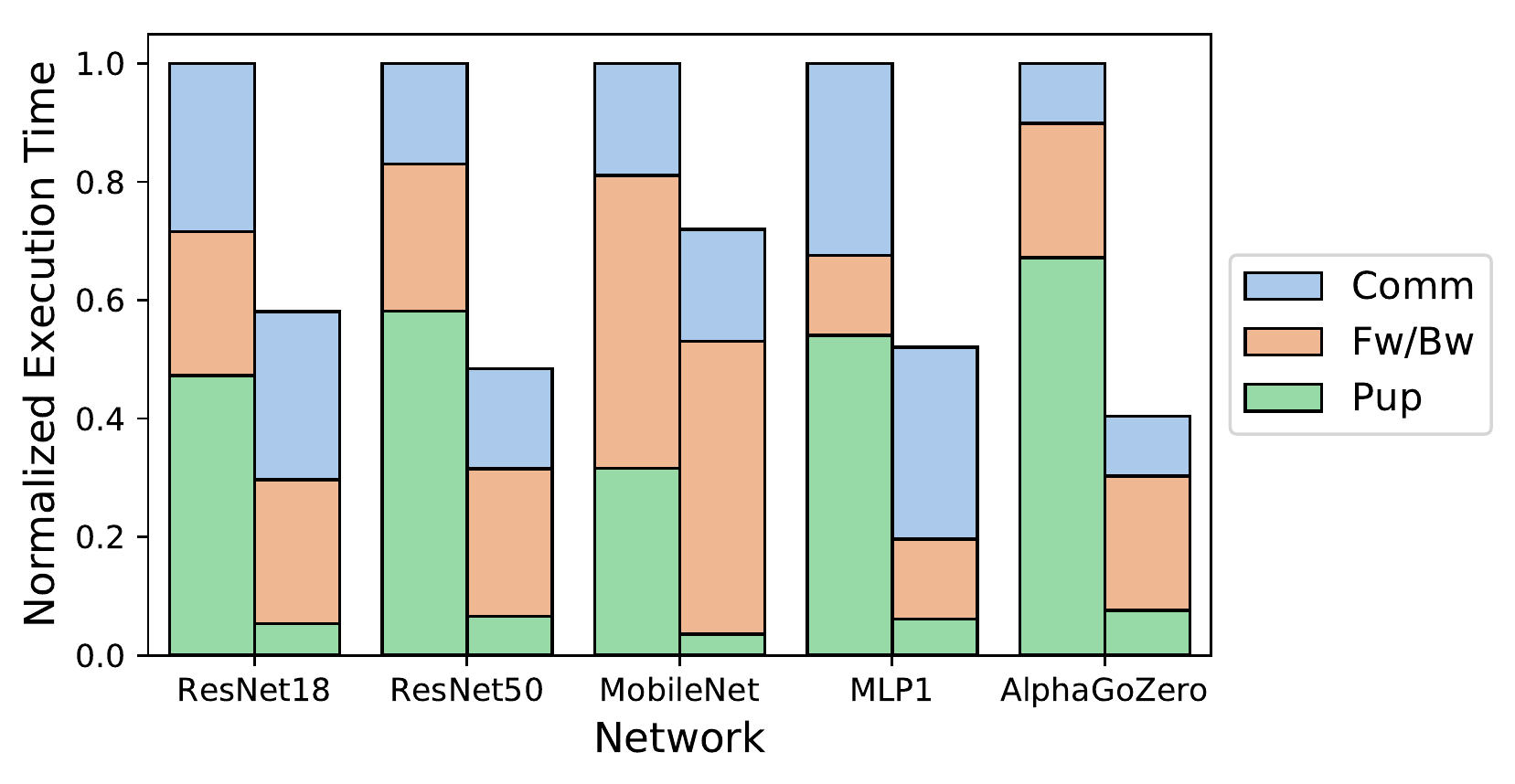}
\vspace{-1mm}
 \caption{Speedups for distributed training.}
\vspace{-1mm}
 \label{fig:DDL}
  \end{figure}

\subsection{Distributed Data Parallelism}
\label{sec:ddl}

To find out the potential for \pimname to work for distributed data parallel distributed learning, we measured the performance gain over the same set of networks, with a modest number of 4 NPU nodes, each taking a quarter of the minibatch as shown in \figurename~\ref{fig:DDL}. \JLr{run the simulation with batch size 8}
We assume the nodes are connected in a torus-like network with 100Gbps connections~\cite{224sec} with HW supporting NI as in \cite{inceptionn}.
The communication time is depicted as Comm. for the baseline and \pimname in the legend.
Due to the smaller effective batch size per node, 
\pimname shows much better scalability compared to the baseline, and the performance is almost 2$\times$ better than the baseline with distributed training.

\section{Related Work}
Placing a logic inside DRAM is not entirely new, and in fact has a long history of work accumulated over a few decades.
Starting from execube~\cite{execube} in the 1990s, there has been much work on integrating processor and memory on a single die~\cite{iram, flexram, smartmem, yukon}, but without commercial success.

The idea of PIM revived in mid-2010s, with the aid of 3D stacking technology~\cite{tesseract, PEI, graphP, graphQ, TopPIM, ndc, pimhetero}.
By placing a logic die underneath a few memory dies, processing logic could be placed together with the memory cells without having to share the same silicon technology of the memory that slows down the processing speed.
Meanwhile, there were attempts to exploit the inherent structure of the existing DDR family to perform a certain class of executions~\cite{rowclone, ambit, bufcmp, computedram, gsdram, redram}.
UpMem~\cite{upmem} has fabricated a processor within a memory die and claims that it can obtain a multifold speedup of the various applications.

%\begin{sloppypar}
DNN as a surging application is another strong driving force towards PIMs. For example, Neurocube~\cite{neurocube} executes DNNs using a 3D stacked memory. 
One major stream of work tries to modify the cell array structure to perform massively parallel DNNs operations within DRAM~\cite{xnorpop, dracc, nndram, drisa, mcdram}, SRAM~\cite{ccache, decsram,xnorsram,xcelram}, and emerging memories~\cite{acdimm, parapim, dima, pinatubo, isaac, mca, prime, floatpim}.
However, these technologies all require changing the cell structure itself, and implementing on top of the legacy technologies is complicated (e.g., DDR4/5 protocol). 

\begin{comment}
Our work can be differentiated in that we do not alter the cell or the bank internal structure. 
Instead, we place a small set of modules along with the peripherals, such that the current efforts towards the high density and efficiency can be preserved while performing parallel computations.
\end{comment}

%\end{sloppypar}
Compared to the solutions above, \pimname is a cheaper, easily realizable solution.
Solutions such as \cite{xnorpop, dracc, nndram, drisa, mcdram} involve re-designing the DRAM core cell array. 
While these are better at performance, they are much harder to realize as a product.
On the other hand, \pimname only alters the datapath at the global I/O gating, requiring only a small amount of change in the circuitry.
We believe approaches similar to \pimname has more potential to be accepted to the industry in the nearer future as they fit more smoothly to the current DRAM standards.

Bank group-level parallelism (BGLP)~\cite{bglp} is another related work worth mentioning since it exhibits a related idea with our proposal. 
While not a processing-in-memory work, it decouples the bank group from the other parts of the DRAM. 
The difference with ours is that 
they use buffers to relax the scheduling restrictions, and increases the utilization of the internal banks.

Our work is close to TensorDIMM~\cite{tensordimm}, which organizes a specialized DIMM targeting gather and reduction to speedup 
embedding lookups and tensor manipulations. 
However, TensorDIMM requires the use of buffer chips to the memory channel, and only utilizes rank internal bandwidth. % and requires adding more ranks for the speedup. 
\pimname can achieve more speedup as shown in Section~\ref{sec:eval} 
by utilizing the bank group level parallelism.

\section{Discussion}
\textbf{Supporting Other Kinds of Parameter Update Algorithms}:
In this paper, we have demonstrated \pimname on SGD with momentum and a weight decay term. 
Some algorithms such as NAG~\cite{nesterov} can be supported with \pimname naturally in the same way. % we have demonstrated.
However, there are algorithms that require more complexity, such as a decaying factor~\cite{adagrad, rmsprop} or second order momentum~\cite{adam}.
These extra values require access to adjacent rows in a bank group concurrently. 
Since momentum SGD does not utilize all banks in a bank group at the same time, there is room for these extra parameters to join the computation. 
\rindex{R7}\rev{In a rare case where the number of variables is too large (four in our setting), 
we can run them in multiple passes.
In the first pass a separate array is allocated for storing intermediate values, and in the next pass, the row with intermediate values can be accessed with the remaining values. 
It would slightly degrade the speedup, but the implementation would not be overly complicated.
}
It would require activating and reading the data multiple times, while causing only a small overhead on the overall performance.

\rindex{R6b}\rev{\textbf{Expandability to other applications}: In this paper, we have focused on DNN parameter update as the target for \pimname. However, there are possibilities for supporting other applications without modifying the \pimname architecture too much. 
In fact, most of the element-wise applications such as table scan~\cite{hana}, histogram construction~\cite{puma} or B-trees~\cite{ntfs}.
It requires change in the ALU of the \pimname unit and more generalized command supports. We believe those can be handled with new DRAM standards~\cite{ddr5}, and leave it as a future work.}

\begin{comment}
\textbf{Using per-bank unit instead of per bank group}: 
\end{comment}

\textbf{Learning Rate Scheduling}:
One could raise question about how to apply learning rate scheduling, as we have assumed a fixed learning rate in this work. 
\pimname can be extended to support varying learning rate. % with slightly more logic added. 
Scaling the values each time by 2 can be easily implemented using a shifter. 
For more complicated scheduling such as cosine~\cite{cosine} or polynomial decay~\cite{polynomial}, we may choose to approximate the decaying function as computing the exact value of them is expensive.  
Another way would be to utilize the mode register and let the NPU provide the new learning rate value, at the expense of some performance overhead.

% \subsection{logic layout}
% - draw the circuit, and how it gets placed within dram (like bufcmp case)

\JL{memo: explain somewhere. the pim still works like a normal dram with DDR commands}

\section{Conclusion}
We have proposed \pimname, a practical PIM design based on the extension of the DDR4 protocol for DNN training.
We have demonstrated that \pimname can speed up the update phase of the DNN training up to around \num{8}$\times$, leading to overall \num{1.94}$\times$ performance gain. % according to the setup.
\pimname poses only a negligible overhead to the DRAM and is fully controlled by the memory controllers.
Even though the proposed design is based on DDR4 SDRAM, we believe similar designs can be adopted to other memories such as HBM, HMC or GDDR. It is expected to show similar speedups or improvement if we exploit more bank group numbers in advanced memory technologies toward high-performance NPUs.

\section*{Acknowledgment}
\small{
This work was supported by 
National Research Foundation of Korea (NRF) grant funded by the Korea government(MSIT) (No.2020R1F1A1074472) %기초
and 
R\&D program of MOTIE/KEIT [No. 10077609, Developing Processor-Memory-Storage Integrated Architecture for Low Power, High Performance Big Data Servers] % 류수정교수님1
and
Institute of Information \& communications Technology Planning \& Evaluation (IITP) grant funded by the Korea government (MSIT) (No.2020-0-01361, Artificial Intelligence Graduate School Program (Yonsei University))  % Yonsei AI
and
Samsung Advanced Institute of Technology, Samsung Electronics Co., Ltd. %류수정교수님2
}

% \balance

% \newpage

%%%%%%% -- PAPER CONTENT ENDS -- %%%%%%%%

%%%%%%%%% -- BIB STYLE AND FILE -- %%%%%%%%
\bibliographystyle{IEEEtranS}
\bibliography{refs}

%\printbibliography

\end{document}